\documentclass{article}

\usepackage[preprint]{neurips_2025}

\usepackage[utf8]{inputenc} 
\usepackage[T1]{fontenc}    
\usepackage{hyperref}       
\usepackage{url}            
\usepackage{booktabs}       
\usepackage{amsfonts}       
\usepackage{nicefrac}       
\usepackage{microtype}      
\usepackage{xcolor}         
\usepackage{amsmath}
\usepackage{algorithm}
\usepackage{algpseudocode}
\usepackage{mathtools}
\usepackage{booktabs}
\usepackage{multirow}
\usepackage{subcaption}
\usepackage{needspace}
\usepackage[pdftex]{graphicx}
\usepackage{wrapfig}
\usepackage{enumitem} 
\definecolor{cornflowerblue}{rgb}{0.39, 0.58, 0.93}

\hypersetup{
    colorlinks=true, 
    linkcolor=cornflowerblue, 
    filecolor=magenta,      
    urlcolor=teal,
    citecolor=cornflowerblue,
    pdftitle={Is your batch size the problem? Revisiting the Adam-SGD gap in language modeling},
    }
\usepackage{amsthm} 
\usepackage{amsmath}
\newtheorem{theorem}{Theorem}

\newcommand{\sign}{\text{sign}}
\newcommand{\E}{\mathbb{E}}

\DeclareMathOperator\erf{erf}
\DeclareMathOperator\diag{Diag}
\definecolor{darkgreen}{rgb}{0.0, 0.5, 0.0}
\definecolor{darkpurple}{rgb}{0.4, 0.0, 0.4}
\definecolor{softdarkgreen}{rgb}{0.0, 0.4, 0.2}   
\definecolor{softdarkpurple}{rgb}{0.4, 0.2, 0.5}  
\title{Is your batch size the problem? Revisiting the Adam-SGD gap in language modeling}

\author{%
  Teodora Srećković\thanks{Correspondence to \texttt{teodorasrec@gmail.com}}, \quad Jonas Geiping, \quad  Antonio Orvieto \\
  Max Planck Institute for Intelligent Systems \\
  ELLIS Institute Tübingen, Tübingen AI Center \\
}

\begin{document}

\maketitle

\begin{abstract}
  Adam is known to perform significantly better than Stochastic Gradient Descent~(SGD) in language models, a phenomenon for which a number of explanations have been proposed. In this work, we revisit this ``optimizer gap'' through a series of comprehensively tuned baseline training runs for language modeling with Transformers. We exhaustively study how momentum, gradient clipping, and batch size affect the gap between SGD and Adam. Our empirical findings show that SGD with momentum can actually perform similarly to Adam in small-batch settings, if tuned correctly. We revisit existing explanations for Adam's advantage, including heavy-tailed class imbalance, directional sharpness, and Hessian heterogeneity, which struggle to directly explain this phenomenon. Towards bridging this gap in our understanding, by analyzing our Transformer training runs and simple quadratic settings inspired by the literature, we provide new insights, driven by stochastic differential equation models, into the role of batch size on the training dynamics.
\end{abstract}

\section{Introduction}

The Adam optimizer~\citep{kingma2014adam} is used pervasively in deep learning, especially when training large language models~(LMs)~\citep{grattafiori2024llama, liu2024deepseek, biderman2023pythia} and vision Transformers~\citep{dosovitskiy2020image,kumar2022fine} at scale. Industrial practice relies on the success of Adam, and thousands of GPU hours every day are spent at large companies using Adam to train their next-generation large language models.



\looseness -1 Even in new sophisticated optimization pipelines looking to dethrone Adam, such as Muon~\citep{jordan2024muon}, most current implementations~\citep{liu2025muon, shah2025practical} rely on plain Adam with weight decay~(AdamW,~\citet{loshchilov2018decoupled}) for critical subsets of parameters, such as normalization layers, text embeddings and prediction heads. This new world is still a bit surprising. Up until around the year 2018, the Adam optimizer was in occasional use, but stochastic gradient descent (SGD) with momentum was known to lead to neural networks with better accuracy on unseen data \citep{wilson2017marginal}, relegating Adam to speed runs and quick comparisons \citep{goyal2017accurate}. Yet, from the start, language modeling with Transformers required Adam. In fact, Transformer LMs have been reportedly untrainable with SGD~\citep{xiong2020layer}, especially due to the critical parameters listed above.


\begin{figure}[tb]
  \centering
  \includegraphics[width=1\linewidth]{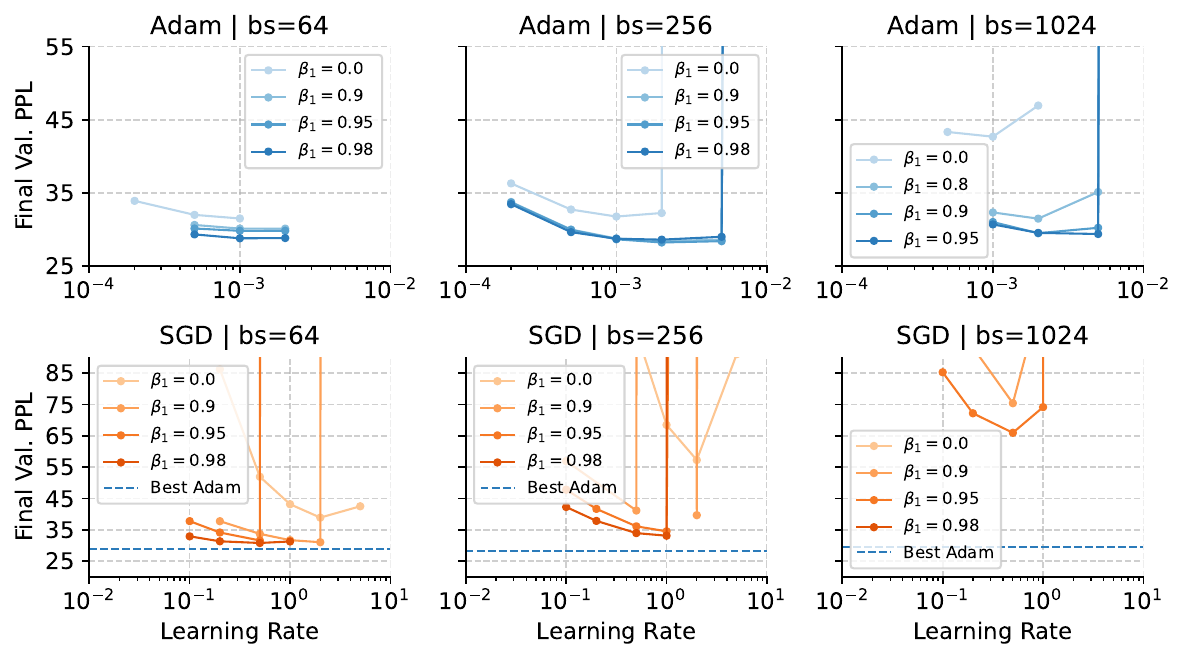}
  \vspace{-.6cm}
  \caption{Learning rate and momentum sweep for SGD and Adam across batch sizes under a fixed compute budget of 1.3B tokens. Perplexities are measured on 100M held-out tokens. All experiments use a 160M-parameter model trained on the SlimPajama dataset with a sequence length 512 and global gradient norm clipping. \textbf{Adam performs consistently across batch sizes, while SGD performs poorly at large batch sizes but gets closer to Adam as batch size decreases}. The dashed blue line in the second row indicates the best-performing Adam configuration for each batch size. Best-performing hyperparameters and perplexities for both optimizers are listed in~\autoref{tab:optimizer_by_bs}.}
  \label{fig:adam_vs_sgd_bs_sweep}
  \vspace{-.2cm}
\end{figure}

Over the years, researchers have offered a number of compelling explanations regarding the remarkable performance of Adam compared to SGD in language modeling, attributing it either to the peculiar noisy nature of text data~\citep{zhang_why_2020, zhang_why_2020-1} or the heterogeneous structure~\citep{noci2022signal, zhang_why_2024} of the Transformer architecture~\citep{vaswani2017attention} --- comprising semantically and structurally dissimilar layers. While most hypotheses regarding the Adam-SGD gap can help guide our understanding~\citep{ahn_linear_2024}, a particularly crucial insight was recently brought to light by~\citet{kunstner_noise_2023}: the Adam-SGD gap is also observable in full-batch training, and is hence clear that the stochastic and potentially heavy-tailed nature of stochastic gradients may not be the challenge Adam is able to tackle.
Inspired by the latter discussion, we take an \textbf{orthogonal approach}:

\begin{center}
    \textit{Instead of asking why Adam often outperforms SGD, we wonder:\\In which Transformer-based language model training setting, if any, does SGD work?}
\end{center}

In other words, while most recent works try to maximize the gap between SGD and Adam in order to explain it more easily, we here try to minimize it -- without sacrificing scale or performance. We believe such a view is novel in the literature and can provide valuable insights into the Adam-SGD gap. In particular, it can help identify settings that falsify existing hypotheses about the gap and enumerate necessary criteria that explanations must fulfill. Our contributions are as follows:


\vspace{-0.2cm}
\setlist[itemize]{leftmargin=1em} 

\begin{itemize}
    \item Despite our own surprise, we show that LMs can be trained with SGD as effectively as Adam at the same token budget, as long as the batch size is chosen small enough, and hyperparameters, such as clipping and momentum, are chosen correctly. We found this holding even at a scale of 1B parameters. While this setting is clearly not convenient in standard pretraining practice on multiple devices, it provides a new lens into understanding the Adam-SGD gap.
    \item To inspect this phenomenon, we carefully revisit prior explanations --- such as heavy-tailed class imbalance~\citep{kunstner_heavy-tailed_2024}, directional sharpness~\citep{pan_toward_2023}, and Hessian heterogeneity~\citep{zhang_why_2024} --- in our setup. While our experiments confirm that these explanations can shed light and are useful to describe settings where Adam outperforms SGD, we find that no prior work can directly explain why SGD can outperform Adam at low batch sizes, while achieving satisfactory performance. Notably, in stark contrast with works attributing the gap to heavy-tail noise, we observe that increased stochasticity actually reduces the Adam–SGD gap.
    \item We enhance our intuition by further studying what makes SGD suboptimal and potentially unstable at high batch sizes. To do this, we present ablations on gradient clipping and learning rate grafting~\citep{agarwal2020disentangling}, and inspect their effect on performance.
    \item Inspired by our observed empirical correspondence between Transformers dynamics and the simplified heterogeneous quadratic setup of~\citep{zhang_why_2024} at different noise levels, we leverage this setup to further study why adaptive optimization may have a different batch size sensitivity compared to SGD. Our analysis is rooted in recent works on SDE models~\citep{malladi_sdes_2023, compagnoni2025adaptive}, and our findings and theoretical connections provide evidence of acceleration for Adam in the large batch settings.
\end{itemize}
\vspace{-2mm}
Together, our findings paint a new picture of the optimizer gap, and suggest that batch size --- and thus the scale and structure of gradient noise --- should be explicitly considered in future analyses. Moreover, our results shift the discussion to considerations on the critical batch size of each optimizer, and can provide practical hints in low-resource and small-scale settings, where small batches are common and optimizer memory usage is critical.

\vspace{-.1cm}
\subsection{Related work}
\vspace{-.2cm}
\label{sec:rw}

\textbf{Class imbalance.} \citet{kunstner_heavy-tailed_2024} explains the advantage of Adam over SGD on language tasks through the heavy-tailed class imbalance in token distributions. They show that the loss for rare tokens decreases significantly more slowly when training with SGD than with Adam, making SGD training inefficient in such settings. In contrast, Adam makes steady progress even on these low-frequency tokens. Their empirical findings show that this explanation is robust to different architectures and settings, suggesting that the performance gap is primarily driven by class imbalance. This explanation is not limited to text data and Transformers: the authors show that an Adam-SGD gap consistently appears in class-imbalanced scenarios, but vanishes when the data is balanced.

\textbf{Transformer architecture.}
Another line of work focuses on the specific characteristics of Transformer architectures. \citet{zhang_why_2024} provide a Hessian-based perspective, showing that Transformers have a block-heterogeneous Hessian spectrum, meaning the Hessian spectrum varies significantly across parameter blocks. In such settings, Adam outperforms SGD by a large margin, while both optimizers perform similarly on architectures with a more homogeneous Hessian. They empirically show that this finding holds across different data modalities and architectures, and that Adam outperforms SGD even on ViT models, differing from the findings of \citet{kunstner_heavy-tailed_2024}. \\
In contrast, \citet{tomihari_understanding_2025} focus on gradient heterogeneity, explaining Hessian heterogeneity as a consequence of the correlation between gradients and the Hessian. They observe that, in Transformers, a large disparity in gradient norms across parameters leads to optimization challenges for SGD, which Adam's adaptivity can address.\\
Finally, through empirical studies, \citet{zhao_deconstructing_2024} find that adaptive optimizers have stable performance over a wide range of hyperparameter settings, while SGD is highly sensitive and often requires careful tuning. They also confirm that full Adam-style adaptivity is not always necessary, showing that their proposed method, which applies adaptivity only to normalization layers and the final layer, can close most of the gap compared to Adam in their setting.

\textbf{Heavy-tailed gradient noise.}
Earlier work by \citet{zhang_why_2020} asks whether the nature of stochastic gradient noise explains why the Adam-SGD gap exists in Transformer models but not in other architectures. They show that Transformer models produce gradient noise with heavy-tailed distributions, in contrast to nearly Gaussian noise in CNNs.  They argue that heavy-tailed gradient noise degrades the performance of SGD, while Adam demonstrates greater robustness.\\
However, evidence from \citet{kunstner_heavy-tailed_2024} shows that noise alone is not the primary cause of Adam's superiority, since the gap exists even in the full-batch setting. They find that the performance gap persists, and even that Adam's advantage grows, as stochastic noise vanishes.

\textbf{Optimization trajectories.}
Several researchers have investigated how Adam differs from SGD by characterizing the path taken during training. \citet{jiang_how_2022} analyze local geometry along training paths and define a statistic that measures the uniformity of the diagonal of the Hessian. On LMs, they find that Adam's trajectory consistently moves through regions where this measure is significantly smaller than the values found along the trajectory of SGD with momentum.\\
Similarly, \citet{pan_toward_2023} introduce directional sharpness as a metric to explain Adam's faster convergence in Transformers. Rather than examining the entire Hessian, they look at the sharpness along the update direction at each step, showing that Adam makes updates in directions with much smaller sharpness than SGD. Although these measures help characterize the training dynamics of SGD and Adam, they appear to correlate with the performance gap rather than fully explain it.


\textbf{Evidence from simplified settings.}
Recent work shows that the Adam-SGD gap persists even in simplified Transformer architectures. \citet{ahn_linear_2024} demonstrate that the characteristic optimization challenges mentioned above also appear in shallow linear Transformers, models without nonlinear activations, on a linear regression task.

\section{Adam vs. SGD: Effects of hyperparameters and training regimes}
\label{sec:adamvssgd}
To systematically investigate the performance gap between Adam and SGD, we conduct a series of experiments in language modeling using a conventional Transformer architecture. Our goal is to understand how this gap evolves under various training regimes and hyperparameter configurations.

\subsection{Experimental setup}
We conduct most of our experiments on the SlimPajama~\citep{cerebras2023slimpajama} dataset using a 160M-parameter nanoGPT~\citep{karpathy2022nanogpt} implementation enhanced by recent advancements. This model has 12 attention layers, a width of 768, and 12 attention heads. MLP blocks use a GLU~\citep{shazeer2020glu} with an expansion ratio of 8/3. The model uses RMSNorm normalization~\citep{zhang2019root} layers and RoPE positional encodings~\citep{su2024roformer}. Full details are provided in the \autoref{app:exp-details}. We also experiment with larger models, up to 1B parameters with a Pythia configuration~\citep{biderman2023pythia}, and on the Fineweb dataset~\citep{penedo2024the}.

\looseness -1 We do not apply weight decay in any of our experiments to eliminate potential side-effects. For both SGD and Adam, we perform global gradient norm clipping on raw gradients~(e.g., before applying momentum) unless otherwise stated. The $\beta_2$ parameter for Adam is fixed at 0.95 throughout all experiments, as common in the literature~\citep{biderman2023pythia}. SGD always refers to SGD with momentum unless explicitly stated otherwise. Other details regarding sequence length, batch size, training budget, and hyperparameter grids are reported directly in the respective sections. 

\subsection{Effect of batch size on the Adam-SGD gap}\label{sec:adamsgdgap}
We first study how the gap between Adam and SGD changes with batch size under a fixed compute budget, when momentum and learning rate are tuned.

\vspace{-3mm}
\paragraph{Setup.} All experiments use a sequence length of 512, a fixed token budget of 1.3B tokens, and a cosine learning rate scheduler~\citep{loshchilov2016sgdr} with 10\% warmup. We compare three batch sizes: 64, 256, and 1024. The learning rate and momentum values are tuned for both optimizers at a batch size of 256. A sweep is performed over 5 learning rates and momentum values of 0.9, 0.95, and 0.98, including runs without momentum. High momentum values are motivated by findings from \citet{zhao_deconstructing_2024}, where SGD performs best with momentum 0.98. Based on the optimal learning rate found at batch size 256, we scale down the learning rate grid for batch size 64 and scale it up for batch size 1024, sweeping over 3 values in each case. Results are reported as the final validation perplexity evaluated on 100M held-out tokens and are shown in \autoref{fig:adam_vs_sgd_bs_sweep}. Some settings become unstable at very large learning rates, where one run may succeed, even if the median run diverges. In those settings, we report runs at the largest stable learning rate as optimal. 

\begin{wrapfigure}[15]{R}{0.45\textwidth}
\vspace{-7mm}
  \centering
  \includegraphics[width=0.45\textwidth]{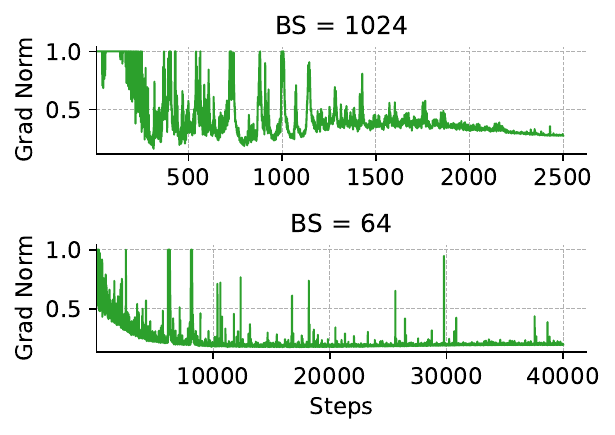}
  \vspace{-5mm}
  \caption{Gradient norm after clipping (threshold 1.0) shows that clipping is more frequent in large-batch training. The setup for these runs is the same as \autoref{fig:adam_vs_sgd_bs_sweep}.}
  \label{fig:clip_norm}
\end{wrapfigure}

\vspace{-3mm}
\paragraph{Results.} Adam shows similar performance across different batch sizes under a fixed token budget, as shown in \autoref{fig:adam_vs_sgd_bs_sweep}. Surprisingly, \emph{SGD can match Adam when training with small batch sizes}, but the gap increases as the batch size grows. For both SGD and Adam, momentum becomes crucial once the batch size is increased, as noted also by \citet{kunstner_noise_2023} and \citet{zhao_deconstructing_2024}. 


We also find that using a relatively small sequence length of 512 is not a crucial factor in these dynamics. As we show in the next section, qualitatively the same behavior can be observed when using a sequence length of 2048 -- as long as the number of tokens per iteration is held constant. This suggests that performance differences can be attributed to the \emph{effective batch size (in tokens)} at a given sequence length, rather than sequence length alone.


\paragraph{Clipping acts differently at different batch sizes.} We observe that gradients are clipped more frequently when training with SGD at large batch sizes, as shown in \autoref{fig:clip_norm}. Additionally, at small batch sizes, SGD performs equally well even without clipping; instead, at large batch sizes, training diverges if clipping is not employed. 

\vspace{-3mm}
\paragraph{Warmup length is not a confounder.} 
We also verify that warm-up length is not a confounding effect, sweeping 5-20\% warmup schedules in our cosine with warmup scheduler. 

\needspace{5\baselineskip}
\subsection{Are large batch sizes the problem, or is it the number of steps?}
\label{sec:big_sweeps}

\begin{figure}[t] 
  \centering
  \includegraphics[width=1\linewidth]{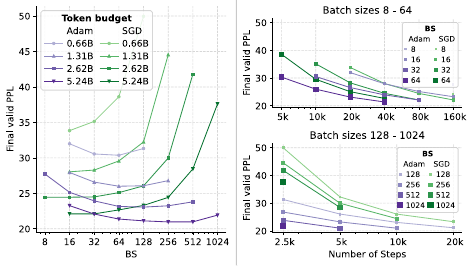}
  \vspace{-.7cm}
  \caption{SGD (green) and Adam (purple) performance across batch sizes. Left: fixed token budget (darker colors – more tokens); the gap increases with batch size across all token budgets. Right: fixed number of steps (darker colors – larger batch sizes); the gap decreases with the number of steps. SGD improves with longer training and can match Adam, given a sufficiently small batch size.}

  \label{fig:steps_tokens_bs}
  \vspace{-.25cm}
\end{figure}

Our previous experiments show that SGD can match Adam in small-batch settings when both optimizers are carefully tuned. Crucially, note that in \autoref{fig:adam_vs_sgd_bs_sweep} all methods see a total of 1.3B tokens. This implies that, e.g., at batch size 1024, methods perform 1/16 of the steps compared to batch size 64. This observation raises an important question: does SGD truly break at large batch sizes, or is it simply slower to converge, compared to Adam, at higher batch sizes? In other words, \textit{can SGD reach Adam-level performance even at higher batch sizes, if given more training steps?}

To investigate this, we compare performance across batch sizes under two training regimes: (1) a fixed token budget and (2) a fixed number of steps. This comparison allows us to disentangle the effects of slow convergence from actual poor optimization performance.

\begin{figure}[b]
  \centering
   \vspace{-.3cm}
  \includegraphics[width=\linewidth]{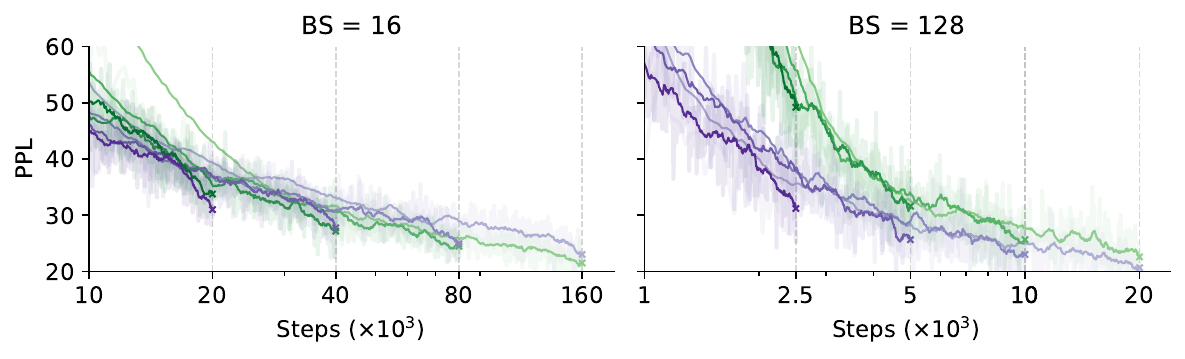}
   \vspace{-.7cm}
  \caption{Perplexity during training for SGD (green) and Adam (purple) across different training lengths in small- and large-batch settings. With high opacity, we plot the filtered~(exponential moving average) PPL values; low opacity lines show the raw values. For both, the gap decreases the longer we train. For a small batch, at the maximal number of steps, SGD even performs better than Adam.} 
  \label{fig:ppl_over_steps}
\end{figure}

\vspace{-3mm}
\paragraph{Setup.} The experimental setting is the same as in the previous section, except with a sequence length of 2048 -- which we increased to ablate on this factor for the second experiment. We train models across a range of batch sizes, from 8 to 1024, and run for different numbers of steps, ranging from 2.5k to 160k. All runs use a fixed warmup of 2000 steps. We switch from the cosine scheduler used previously to a WSD scheduler \citep{hagele_scaling_2024}, to better compare runs before learning rate decay begins. We limit the total token budget between approximately 650M and 5.2B tokens: models using larger batches are not trained for the largest number of steps, while models using smallest batches are trained only for a large number of steps. For SGD, we choose a momentum $\beta=0.98$ and run three distinct learning rates: $1, 0.5$ and $0.25$. For Adam, we choose $\beta_1=\beta_2=0.95$ as suggested by modern literature~\citep{zhao_deconstructing_2024, zhang_how_2025} and report performance for the best performing learning rate in the grid $[1e-3,2e-3,4e-3]$. Both the SGD and Adam configurations are suggested from our more careful tuning performed in \autoref{fig:adam_vs_sgd_bs_sweep}.

\vspace{-2mm}
\paragraph{Results.} The left panel of \autoref{fig:steps_tokens_bs} clearly shows that, at a fixed token budget, Adam improves with larger batch sizes, up to some critical batch size. In contrast, SGD shows a drastically opposite trend --- performance consistently degrades as batch size increases. Under a fixed token budget, matching performance between Adam and SGD is conditional on using very small batch sizes, leading to significantly longer training and poor memory usage. This result highlights a key limitation of SGD: it is highly inefficient in realistic model training, where large batches are required for efficiency. 

In the right panel on \autoref{fig:steps_tokens_bs}, we show performance after training with various numbers of steps. The gap between Adam and SGD grows with batch size, but SGD improves significantly with more steps and can eventually match or even outperform Adam with long enough training. This illustrates that SGD is not necessarily bad, just very slow to converge in large-batch settings. Or to, put it differently: While Adam is able to accelerate (in terms of progress per step) with increased batch size, SGD cannot --- its \textbf{critical batch size} is close to 1.

In addition to \autoref{fig:steps_tokens_bs}, we report the perplexity during training for SGD and Adam with batch sizes 16 and 128 in \autoref{fig:ppl_over_steps}. For both optimizers, the gap decreases as training progresses. In the small-batch setting, SGD even outperforms Adam at the maximum number of steps. We show the same plots for other batch sizes in \autoref{app:add-res}.

\vspace{-2mm}
\paragraph{Scaling experiments.} 
To test whether our findings persist at scale and across datasets, we experiment with larger models (250M, 410M, and 1B parameters) and include additional training on the FineWeb dataset~\citep{penedo2024the} . We repeat the same experiments, varying token budgets and number of training steps, for the 160M model on SlimPajama and 250M model on FineWeb. Results and setup details are reported in \autoref{app:exp-details} and \autoref{app:add-res}, showing that our core claims hold when scaling up the model and switching to a different dataset.

To further test whether SGD can outperform Adam at scale, we train:
\vspace{-0.2cm}
\begin{itemize}
    \item \textbf{410M} model on SlimPajama (sequence length 2048, batch size 8, 500k steps);
    \item \textbf{1B} model on FineWeb (sequence length 1024, batch size 16, steps 850k).
\end{itemize}
\vspace{-0.2cm}

\begin{figure}[h]
  \centering
  \begin{subfigure}[t]{0.48\textwidth}
    \centering
    \includegraphics[width=\linewidth]{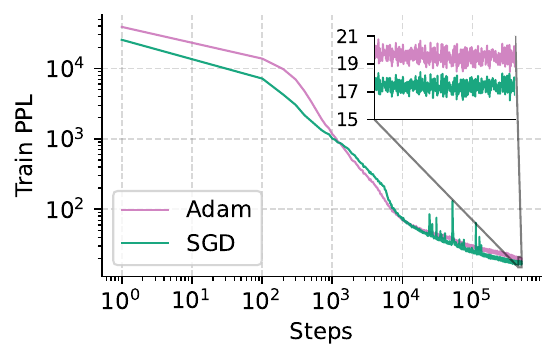}
    \caption{410M model on SlimPajama (seq. length 2048, batch size 8, 500k steps) -- 1.5 days of training.}
    \label{fig:trajectory_410M}
  \end{subfigure}
  \hfill
  \begin{subfigure}[t]{0.48\textwidth}
    \centering
    \includegraphics[width=1\linewidth]{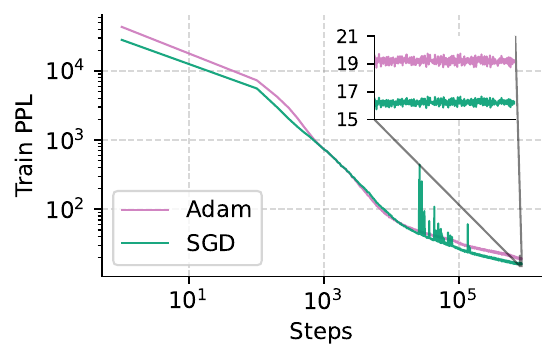}
    \caption{1B model on FineWeb (seq. length 1024, batch size 16, 850k steps) -- 5 days of training.}
    \label{fig:trajectory_1B}
  \end{subfigure}
  \caption{SGD can outperform Adam even at 410M and 1B scales in small-batch regimes.}
  \label{fig:trajectory_large_models}
\end{figure}

Full training details and learning rate tuning plots are provided in \autoref{app:exp-details}. Trajectories for both models are shown in \autoref{fig:trajectory_large_models}, demonstrating that \textbf{SGD can outperform Adam even at a 410M and 1B scale}. For these experiments, we choose the largest batch size that can fit in our NVIDIA A100 80GB card, and do not use gradient accumulation.


\section{Revisiting prior explanations}
\label{sec:prior_exp}
\looseness -1 Several recent works have proposed explanations for Adam's advantage over SGD through the lens of data or architectural properties~(see~\autoref{sec:rw}). All these explanations improve our understanding of the performance gap, yet most are restricted to specific scenarios where the gap between Adam and SGD is pronounced. In this section, we revisit these explanations through the lens of our findings in \autoref{sec:adamvssgd} and ask: \textit{Can they also account for strong SGD performance in small-batch settings?} 

We found that, while no discussion in the literature can properly account for the empirical evidence we presented, the heterogeneous toy quadratic example of~\citet{zhang_why_2024} offers particularly valuable insights, which we develop in \autoref{sec:theory}.

\subsection{Heavy-tailed class imbalance}

\begin{wrapfigure}[14]{R}{0.45\textwidth}
\vspace{-10mm}
  \centering
  \includegraphics[width=0.45\textwidth]{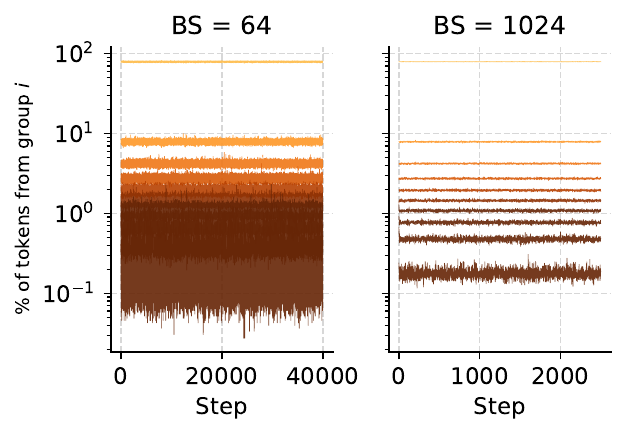}
  \vspace{-6mm}
  \caption{Batch token distribution for batch sizes 64 and 1024. Lighter colors show less frequent tokens. Statistics at lower batch sizes are noisier but of similar magnitude.}
  \label{fig:imbalance_cnt}
\end{wrapfigure}

Prior work by \citet{kunstner_noise_2023} attributes Adam’s advantage over SGD to heavy-tailed class imbalance in token distributions, showing that SGD has difficulty optimizing rare (least common) tokens. We follow their methodology and group all tokens from the training set into 10 frequency groups, from the first group, which contains the 10\% least frequent tokens, to the last group, which contains the 10\% most frequent ones. 

We apply this analysis to the setting from \autoref{sec:adamsgdgap}, comparing batch sizes 64 and 1024, where SGD performs drastically differently, using runs with the optimal combination of $\beta_1$ and learning rate for each case. We find that class imbalance exists in both cases: the persistence of low- and high-frequency tokens is similar, as shown in \autoref{fig:imbalance_cnt}. However, this does not appear to cause problems for small-batch SGD, suggesting that \textit{class imbalance alone does not imply an Adam-SGD gap across all training regimes}.

\begin{figure}[ht]
  \centering
  \includegraphics[width=0.9\linewidth]{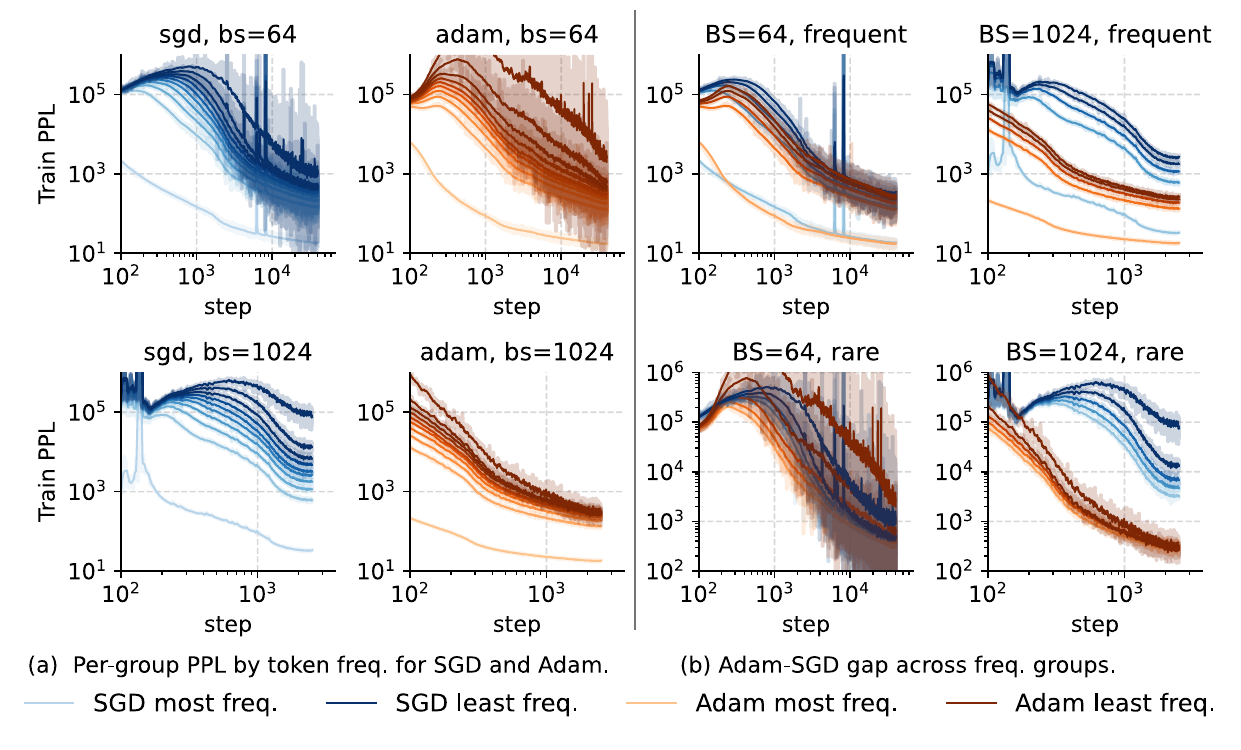}
   \vspace{-.35cm}
  \caption{\textbf{(left)} Perplexity during training for Adam and SGD in small- and large-batch settings, computed per frequency group. SGD shows a larger gap across groups in the large-batch setting, while the opposite holds for Adam. \textbf{(right)} Comparison of the Adam–SGD gap across frequency groups, with more frequent token groups shown on the top row and less frequent ones on the bottom. SGD underperforms Adam across all groups in large batches, with a notably larger gap on rare tokens. This effect is not present in the small-batch setting.}
 \vspace{-.25cm}
  \label{fig:imabalance_ppl}
\end{figure}

We further compute perplexity separately for each frequency group and report it over training. From \autoref{fig:imabalance_ppl} (a), we observe that both optimizers make faster progress on more frequent tokens in all settings, as expected. The relative difference in perplexity between frequency groups is more significant for SGD in the large-batch setting than for the small, while the opposite holds for Adam.

Comparing Adam and SGD across frequency groups in \autoref{fig:imbalance_cnt}, we observe that in the large-batch setting, SGD underperforms Adam across all groups, as shown in \autoref{fig:imabalance_ppl} (b). However, the gap is notably more significant for less frequent tokens, which aligns well with findings from \citet{kunstner_heavy-tailed_2024}, suggesting that rare tokens could be more challenging for SGD in imbalanced settings. In contrast, this effect is not observed for the small-batch regime in our setting, as also clear from the results in~\autoref{sec:adamvssgd}. We would expect this problem of SGD to hold, independent of batch size, but in the setting where SGD works well, the issue disappears.


\subsection{Directional sharpness}

\begin{figure}[h]
  \centering
  \includegraphics[width=\linewidth]{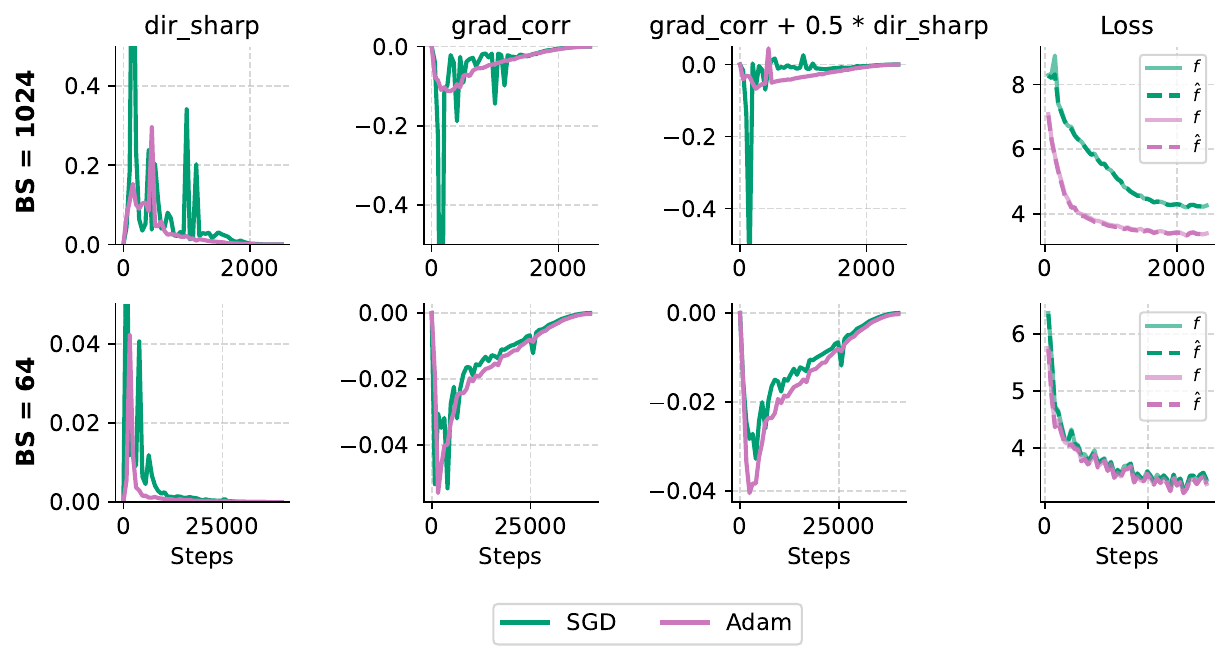}
   \vspace{-.5cm}
  \caption{Gradient correlation, directional sharpness, their sum and second-order loss approximation during training, under small- and large-batch settings. Irrespective of the optimizer, good training trajectories have strong negative gradient correlation and low directional sharpness.}
  \label{fig:dir_sharp}
  \vspace{-.2cm}
\end{figure}

\citet{pan_toward_2023} introduce directional sharpness to explain the optimizer gap by studying a second-order Taylor expansion of the loss along the update direction. In this view, the first-order term (gradient correlation) measures how well the update aligns with the negative gradient, while the second-order term (directional sharpness) measures curvature along that direction. Making optimization progress requires a strong negative gradient correlation and low directional sharpness. Let $f$ be a generic loss to optimize and $x_k$ denote the model parameters at iteration $k$, then
\begin{equation}
    f(x_{k+1}) = f(x_k) + \underbrace{\nabla f(x_k)^\top (x_{k+1} - x_k)}_{\text{gradient correlation}} + \frac{1}{2}\underbrace{(x_{k+1} - x_k)^\top \nabla^2 f(x_k)(x_{k+1} - x_k)}_{\text{directional sharpness}} + O(\eta^3)
    \label{eq:taylor}
\end{equation}
\looseness -1 In \autoref{fig:dir_sharp}, we visualize gradient correlation, directional sharpness, and their sum --- a second-order approximation of loss change, to indicate progress. As in our previous analysis, we compare two settings with drastic performance differences: batch sizes 64 and 1024 from \autoref{sec:adamsgdgap}. In the large-batch setting, SGD shows low gradient correlation and high directional sharpness, resulting in weak or even positive total loss change, as reflected in the sum. In contrast, Adam has higher gradient alignment and lower directional sharpness throughout training. When SGD succeeds, its gradient correlation and directional sharpness closely match Adam’s, producing a negative loss change in the sum. While these metrics align with SGD's success or failure, they do not directly explain why Adam outperforms SGD, nor why SGD performs well in small-batch regimes.

\subsection{Hessian heterogeneity}

From the line of work focusing on the architectural properties of Transformers, \citet{zhang_why_2024} argue that the block-wise heterogeneity of the Hessian spectrum is a key factor behind Adam’s strong performance and the weakness of SGD. They propose that, based on the Hessian structure at initialization, it is possible to predict whether SGD will perform well, \textbf{offering an explanation that is invariant to batch size}. To further explore the effect of batch size on heterogeneous problems, we revisit the simplified quadratic setting from their work and extend it by including batch size variation. We compare optimization on problems with homogeneous~(CNN-like) and heterogeneous~(Transformer-like) Hessians, where both share the same eigenvalue spectrum. We train with SGD and Adam using a cosine learning rate schedule and no clipping. 

\begin{figure}[h]
  \centering
  \includegraphics[width=0.99\linewidth]{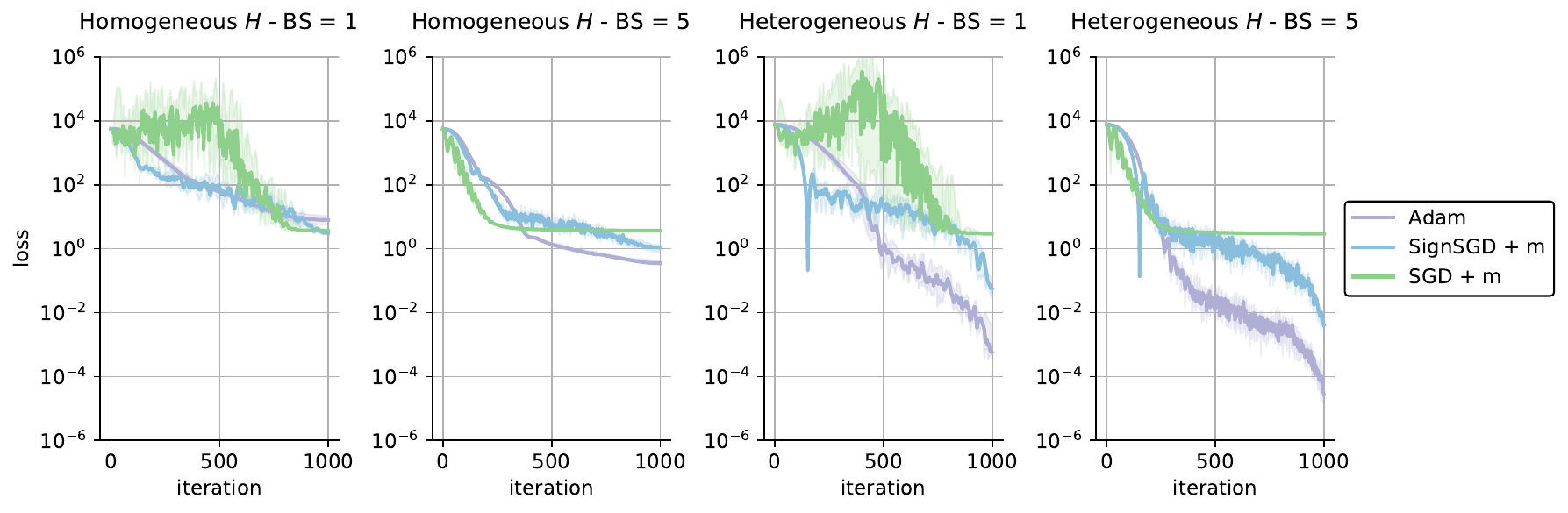}
  \vspace{-.1cm}
  \caption{The gap between Adam and SGD relative to batch size also appears when studying only noisy quadratic models in~\cite{zhang_why_2024}. The Heterogeneous setting is inspired by the Transformer structure, while the Homogeneous by the CNN structure. See the appendix for details. We note that the advantage of Adam of a big batch is also noticeable in the Homogeneous setting, yet much more drastically in the Heterogeneous setting. Learning rates are tuned so to give similar performance at Homogeneous batch size 1. Shown is mean and 2-sigma standard dev. for 10 runs.}
  \label{fig:quadratic}
\end{figure}

\looseness -1 We observe the following in~\autoref{fig:quadratic}. Details of the setting are provided in \autoref{app:quadratic_details}.
\begin{itemize}
    \item Across batch sizes, the largest Adam-SGD gap is observed in the heterogeneous setting -- this is the result by~\citet{zhang_why_2024}. As noted by~\citep{kunstner_noise_2023}, a similar pattern can be observed for signed momentum~\citep{bernstein_signsgd_2018}. We develop on this in \autoref{sec:theory}.
    \needspace{5\baselineskip}
    \item Adam benefits from increasing batch size, in both homogeneous and heterogeneous Hessian problems. SGD does not seem to profit from an increased batch size. We motivate this theoretically in \autoref{sec:theory}: early progress in SGD performance is driven by number of iterations, while it is batch size dependent for signed gradient methods. 
\end{itemize}

To summarize, a performance \textbf{boost can be observed} at higher batches for both SignSGD and Adam, \textbf{regardless of heterogeneity}. While we confirm that heterogeneity amplifies the gap between SGD and adaptive methods, this result showcases that the phenomenon we study in this paper may not be limited to the heterogeneous setting. This is a key insight, bringing the discussion back to a statistical level where the landscape structure plays a less crucial role. We develop on this in ~\autoref{sec:theory}.

\section{Understanding how performance relates to batch size}

In \autoref{sec:prior_exp}, we saw that while prior work sheds light on the Adam-SGD gap in the large-batch regime, it remains unclear how batch size itself factors into these explanations.

Towards gaining more insights, we proceed as follows:
\begin{itemize}
    \item In \autoref{sec:grafting} and \autoref{sec:clipping} we approach this from the SGD angle: \textit{What goes wrong~(see e.g. \autoref{fig:clip_norm}) for SGD in large-batch settings that does not appear at small batch sizes?} To investigate this, we attempt to separate which component of the SGD update is more problematic in settings where it fails --- is it its direction or magnitude?. We focus on the setting from \autoref{sec:adamsgdgap}, using batch size 1024 and the optimal combination of $\beta_1$ and learning rate.
    \item In \autoref{sec:theory} we take a different approach, one based on noise statistics and adaptivity in a setup which is non-specific to the Transformer architecture. This analysis is inspired by the results in \autoref{fig:quadratic}, showing how adaptive methods may profit from large batch sizes regardless of the Hessian structure. Using theoretical tools, we prove here that while SGD performance in early training is dominated by number of iterations, the dynamics of signed momentum methods~(cf. \autoref{fig:quadratic}) showcase a strong dependency on batch size from the very first iterations.
\end{itemize}

\subsection{Insights from Grafting}
\label{sec:grafting}
\begin{figure}[tb]
  \centering
  \begin{minipage}[t]{0.48\textwidth}
    \centering
    \includegraphics[width=0.85\linewidth]{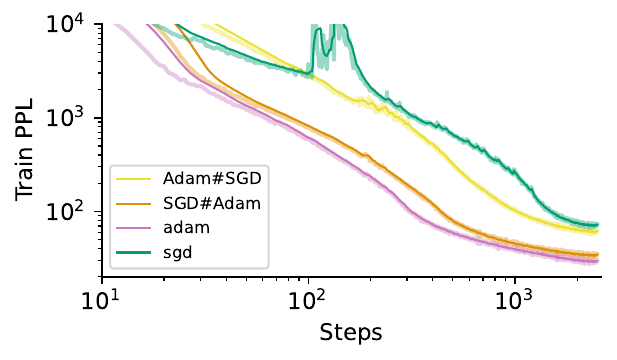}
    \vspace{-0.25cm}
    \caption{Grafting in large-batch training: using Adam’s direction results in performance closer to Adam, while SGD direction leads to results closer to SGD.}
    \label{fig:grafting}
  \end{minipage}%
  \hfill
  \begin{minipage}[t]{0.48\textwidth}
    \centering
    \includegraphics[width=0.85\linewidth]{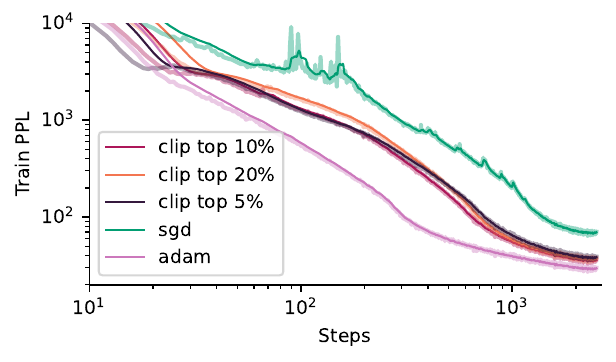}
    \vspace{-.25cm}
    \caption{Adaptive clipping with different percentages of clipped coordinates in large-batch training. It improves SGD but still does not fully match Adam.}
    \label{fig:clipping}
  \end{minipage}
  \vspace{-2mm}
\end{figure}

To isolate the role of update direction and magnitude, we use the grafting technique proposed by  \citet{agarwal2020disentangling}, which applies the update direction of one optimizer with the magnitude of another. We train the model in the large batch setting, using both combinations: 1) SGD magnitude with Adam direction (SGD\#Adam), and 2) Adam magnitude with SGD direction (Adam\#SGD). We use the optimal $\beta_1$ from \autoref{sec:adamsgdgap}, and sweep the learning rate for the grafted update. We report the training perplexity using the optimal learning rate for both grafting combinations in \autoref{fig:grafting}. As shown, using SGD magnitude with Adam's direction performs comparably to Adam, while the reverse combination behaves similarly to SGD. This suggests that the \emph{update direction} is the more problematic component of the SGD update in large-batch training.

\subsection{Insights from Adaptive Clipping}
\label{sec:clipping}

\begin{wrapfigure}[19]{R}{0.5\textwidth}
  \centering
  \vspace{-1.3cm}
  \includegraphics[width=1\linewidth]{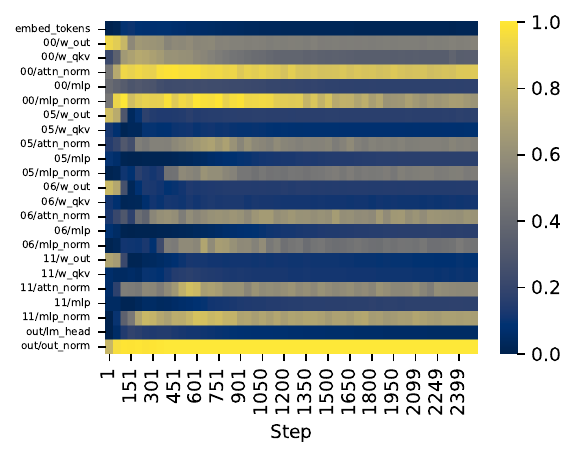}
  \vspace{-0.6cm}
  \caption{Fraction of clipped momentum coordinates per layer during training, using $p = 10\%$ adaptive clipping. Only a subset of blocks is shown for clarity, as similar patterns are observed across all blocks. Clipping is present across all parameters, but most pronounced in normalization layers.}
  \label{fig:clipping_layers}
\end{wrapfigure}

The perspective that direction is the core problem aligns with the observation that global norm clipping does not help much in large-batch training with SGD. If the direction is the main issue, simply rescaling the gradient norm does not lead to better updates.

\looseness -1 To investigate this further, we experiment with adaptive clipping, motivated by \citet{pan_toward_2023}. As shown in \autoref{alg:adapt_clip}~(Appendix), we clip the top $p$\% of the largest momentum coordinates at each step. We test several values of $p$ (5, 10, and 20 \%). For each value, we keep the optimal $\beta_1$ from the previous setting and tune the learning rate. Clipping with $p = 10\%$ performs best, but we observe that performance does not vary much across different values of $p$. This method helps reduce the gap between SGD and Adam, as shown in \autoref{fig:clipping}. This suggests that a subset of larger update coordinates consistently contributes to poor update directions and slows down SGD in large-batch training. In contrast, small-batch training does not present the same problematic coordinates.


Further, we ask whether certain groups of parameters are more likely to produce problematic coordinates. To explore this, we inspect which layers the clipped momentum coordinates come from, using the best-performing setting with $p=10\%$. In \autoref{fig:clipping_layers}, we show the fraction of parameters within each layer that are clipped, relative to the total number of parameters in that layer. We find that normalization layers are clipped the most, which aligns with findings from \citet{zhao_deconstructing_2024} and \citet{tomihari_understanding_2025}. However, this does not imply that only normalization layers are problematic. As we observe significant clipping across other layers as well, this suggests that large coordinates persist across all parameters, though they are most pronounced in normalization layers.

\subsection{Theoretical insights} 
\label{sec:theory}
Towards explaining the phenomena observed in this paper --- and specifically the quadratic example in \autoref{fig:quadratic} --- we provide an analysis based on results around the interaction between noise and adaptivity in~\citep{compagnoni2025adaptive}. The discussion below is rooted on the observed similarity between behaviors of Adam and SignSGD in \autoref{fig:quadratic}, as well as recent literature on their relation~\citep{kunstner_noise_2023, jordan2024muon}.

Let $X$ denote the model parameters and $\gamma$ denote a batch of size $B$. We denote the stochastic gradient as $\nabla f_{\gamma}(x) := \frac{1}{B} \sum_{i\in \gamma} \nabla (f_i(x))$ and by $\Sigma(x)$ the noise covariance at batch size 1. The stochastic differential equation~(SDE) approximation of SGD reads~\citep{mil1986weak,liu2018diffusion}
{\small
\begin{align}
\label{eq:SGD_SDE}
d X_t = {\color{softdarkgreen}- \nabla f(X_t) dt }+ \sqrt{\frac{\eta\Sigma}{B}}  d W_t,
\end{align}}

We now state a recent result showing that the drift of signed updates -- driving performance in early training -- has an extra dependency on the batch size. A proof sketch is provided below.
\begin{theorem}[\citep{compagnoni2025adaptive}] 
\label{monzio}
Under the assumption of i.i.d. Gaussian noise with diagonal covariance~(and, with minor modifications, for other noise structures and non-diagonal covariance), the following SDE provides a $1$-weak approximation~\citep{mil1986weak} of SignSGD:
{\small
\begin{align}
\label{eq:SignSGD_SDE_Simplified_Insights}
d X_t ={\color{softdarkpurple} - \erf \left(\sqrt{\frac{B}{2}} \Sigma^{-\frac{1}{2}} \nabla f(X_t) \right) dt} + \sqrt{\eta} \sqrt{I_d -\diag \left( \erf \left(\frac{\sqrt{B}\Sigma^{-\frac{1}{2}} \nabla f(X_t)}{\sqrt{2}} \right) \right)^2} d W_t,
\end{align}}
where the error function $\erf(x):=\frac{2}{\sqrt{\pi}} \int_0^x e^{-t^2} dt$ and the square are applied component-wise.
\end{theorem}
While~\citet{compagnoni2025adaptive} provide a similar result for the Heavy-tail noise setting, the Gaussian case already highlights a crucial distinction between signed gradient methods and classical SGD.

\begin{wrapfigure}[17]{R}{0.5\textwidth}
  \centering
  \vspace{-8mm}
  \includegraphics[width=0.9\linewidth]{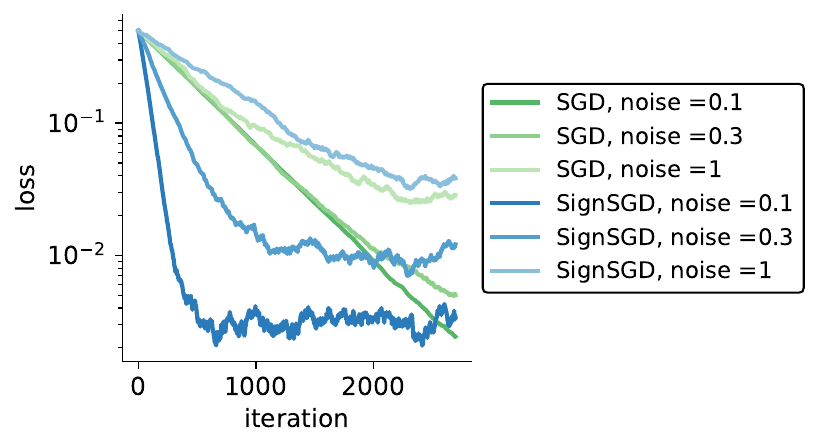}
  \vspace{-2mm}
  \caption{Illustration of theory presented in the section. Optimizing $f(x) = \frac{1}{2}\|x\|^2$, $x\in\mathbb{R}^{100}$. All methods use same learning rate of $1e-3$ and no momentum. Shown is performance under different injected Gaussian noise standard deviation. SGD in early training is dominated by the drift component, which is independent of noise -- progress is driven by number of iterations. For SignSGD, noise~(hence batch size in the more general case) directly affects drift and early progress.}
  \label{fig:theory}
\end{wrapfigure}
\paragraph{Takeaway.} Recall that $\erf$ is linear on a large interval around zero. The local update of parameters is then driven by ${\color{softdarkpurple}- \erf \left(\sqrt{\frac{B}{2}} \Sigma^{-\frac{1}{2}} \nabla f(X_t) \right) dt}$ in the signSGD case, while in the SGD setting, this term is simply ${\color{softdarkgreen}- \nabla f(X_t) dt}$. When everything else is kept constant, increasing the batch size $B$, increases the drift in the direction of the negative gradient by $\sqrt{B}$, up until the saturation point of $\erf$, i.e. a critical batch size.

This analysis provides  evidence for our results: \textit{using large batch sizes accelerates convergence}~(larger drift) in signSGD (and likely also in closely-related algorithms, like Adam), while the performance of SGD in early training is batch-size agnostic and hence driven by the number of iterations~(see \autoref{fig:theory}).

\paragraph{Intuition behind the proof in~\citep{compagnoni2025adaptive}.} We want to study the quantity
$$\sign(m(x)),\qquad m(x)\sim \mathcal{N}(\nabla f(x),\sigma^2/B).$$
That is, $m(x)$ is an estimate of the gradient, which we assume for simplicity to have Gaussian distribution centered around the full-batch gradient $\nabla f(x)$. In expectation, we get~(coordinatewise)
\begin{align*}
    &\E[\sign(m(x))] \\ &= \mathbb{P}[\sign(m(x)) = \sign(\nabla f(x))] \cdot\sign(\nabla f(x)) - \mathbb{P}[\sign(m(x)) \ne \sign(\nabla f(x))]\cdot \sign(\nabla f(x))\\
&= (2\mathbb{P}[\sign(m(x)) = \sign(\nabla f(x))]-1)\cdot \sign(\nabla f(x))
\end{align*}
At this point, the $\erf$ function comes in. Recall that, if $Z\sim \mathcal{N}(\mu,\varsigma^2)$, then if $\ell>\mu$, we have $\mathbb{P}[Z\le \ell] = \frac{1}{2}+\frac{1}{2}\erf\left(\frac{\ell - \mu}{\sqrt{2\varsigma^2}}\right)$. Note that this implies, for $\mu\ge 0$, $\mathbb{P}[Z\ge 0] = \mathbb{P}[\sign(Z) = \sign(\mu)] = \frac{1}{2}+\frac{1}{2}\erf\left(\frac{\mu}{\sqrt{2\varsigma^2}}\right)$. Hence, if $\nabla f(x)>0$ for a specific coordinate, $\mathbb{P}[\sign(m(x)) = \sign(\nabla f(x))] =  \frac{1}{2}+\frac{1}{2}\erf\left(\sqrt{B}\frac{\mu}{\sqrt{2\sigma^2}}\right)$, relative to that coordinate. By symmetry of this argument for negative $\nabla f(x)$, we get exactly the drift term in \autoref{monzio}, discussed above:
$$\E[\sign(m(x))] = \erf\left(\sqrt{\frac{B}{2}}(\sigma^2)^{-\frac{1}{2}}\nabla f(x)\right).$$
Finally, note that Gaussianity is not strictly needed for our insights on batch-size acceleration to hold. As discussed by~\citet{compagnoni2025adaptive} and clear from the argument above on the cumulative distribution, a similar expression can hold even for distributions with heavier tails, such as the $t$-student~(see Corollary C.10 in~\citet{compagnoni2025adaptive}).

\section{Discussion}
Is it impossible to train language models with SGD? In this work, we show that there exist settings, namely small batch sizes with carefully tuned momentum and clipping, where SGD is competitive, even for GPT-2-scale language models. These findings are interesting in their own right for small-scale training runs, for example, on commodity GPUs, where memory is limited. Yet, these findings also crucially inform the space of possible theories for the optimizer gap between Adam and SGD. We revisit a number of promising theories from the literature based on our findings and find that they have limited explanatory power. We argue that, instead, the effect of batch size is a symptom of the importance of gradient noise for this question, and discuss a stronger explanation based on SDEs.

We believe that future theoretical work might be able to better explain the Adam-SGD gap as a function of the batch size. A promising direction is recent work on $(L_0,L_1)$-smoothness~\citep{zhang_why_2020-1}, specifically in the context of signed gradient descent~\citep{compagnoni2025interaction}, as well as $\ell_{\infty}$ geometry~\citep{xie2024implicit, xie2024adam} of transformer models.

\section*{Acknowledgements}
The authors thank Weronika Ormaniec and 
Sajad Movahedi for their insightful comments on our work, and Enea Monzio Compagnoni for his comments and support in the theoretical section. The authors acknowledge the financial support of the Hector Foundation.

\clearpage
\newpage

\bibliographystyle{abbrvnat}
\bibliography{main}

\begin{thebibliography}{51}
\providecommand{\natexlab}[1]{#1}
\providecommand{\url}[1]{\texttt{#1}}
\expandafter\ifx\csname urlstyle\endcsname\relax
  \providecommand{\doi}[1]{doi: #1}\else
  \providecommand{\doi}{doi: \begingroup \urlstyle{rm}\Url}\fi

\bibitem[Agarwal et~al.(2020)Agarwal, Anil, Hazan, Koren, and Zhang]{agarwal2020disentangling}
N.~Agarwal, R.~Anil, E.~Hazan, T.~Koren, and C.~Zhang.
\newblock Disentangling adaptive gradient methods from learning rates.
\newblock \emph{arXiv preprint arXiv:2002.11803}, 2020.

\bibitem[Ahn et~al.(2024)Ahn, Cheng, Song, Yun, Jadbabaie, and Sra]{ahn_linear_2024}
K.~Ahn, X.~Cheng, M.~Song, C.~Yun, A.~Jadbabaie, and S.~Sra.
\newblock Linear attention is (maybe) all you need (to understand transformer optimization), Mar. 2024.
\newblock URL \url{http://arxiv.org/abs/2310.01082}.
\newblock arXiv:2310.01082 [cs, math].

\bibitem[Bernstein et~al.(2018)Bernstein, Wang, Azizzadenesheli, and Anandkumar]{bernstein_signsgd_2018}
J.~Bernstein, Y.-X. Wang, K.~Azizzadenesheli, and A.~Anandkumar.
\newblock {signSGD}: {Compressed} {Optimisation} for {Non}-{Convex} {Problems}, Aug. 2018.
\newblock URL \url{http://arxiv.org/abs/1802.04434}.
\newblock arXiv:1802.04434 [cs, math].

\bibitem[Biderman et~al.(2023)Biderman, Schoelkopf, Anthony, Bradley, O’Brien, Hallahan, Khan, Purohit, Prashanth, Raff, et~al.]{biderman2023pythia}
S.~Biderman, H.~Schoelkopf, Q.~G. Anthony, H.~Bradley, K.~O’Brien, E.~Hallahan, M.~A. Khan, S.~Purohit, U.~S. Prashanth, E.~Raff, et~al.
\newblock Pythia: A suite for analyzing large language models across training and scaling.
\newblock In \emph{ICML}, 2023.

\bibitem[Black et~al.(2022)Black, Biderman, Hallahan, Anthony, Gao, Golding, He, Leahy, McDonell, Phang, et~al.]{black2022gpt}
S.~Black, S.~Biderman, E.~Hallahan, Q.~Anthony, L.~Gao, L.~Golding, H.~He, C.~Leahy, K.~McDonell, J.~Phang, et~al.
\newblock Gpt-neox-20b: An open-source autoregressive language model.
\newblock \emph{arXiv preprint arXiv:2204.06745}, 2022.

\bibitem[Chowdhery et~al.(2023)Chowdhery, Narang, Devlin, Bosma, Mishra, Roberts, Barham, Chung, Sutton, Gehrmann, et~al.]{chowdhery2023palm}
A.~Chowdhery, S.~Narang, J.~Devlin, M.~Bosma, G.~Mishra, A.~Roberts, P.~Barham, H.~W. Chung, C.~Sutton, S.~Gehrmann, et~al.
\newblock Palm: Scaling language modeling with pathways.
\newblock \emph{Journal of Machine Learning Research}, 24\penalty0 (240):\penalty0 1--113, 2023.

\bibitem[Compagnoni et~al.(2025{\natexlab{a}})Compagnoni, Islamov, Orvieto, and Gorbunov]{compagnoni2025interaction}
E.~M. Compagnoni, R.~Islamov, A.~Orvieto, and E.~Gorbunov.
\newblock On the interaction of noise, compression role, and adaptivity under $(l\_0, l\_1) $-smoothness: An sde-based approach.
\newblock \emph{arXiv preprint arXiv:2506.00181}, 2025{\natexlab{a}}.

\bibitem[Compagnoni et~al.(2025{\natexlab{b}})Compagnoni, Liu, Islamov, Proske, Orvieto, and Lucchi]{compagnoni2025adaptive}
E.~M. Compagnoni, T.~Liu, R.~Islamov, F.~N. Proske, A.~Orvieto, and A.~Lucchi.
\newblock Adaptive methods through the lens of {SDE}s: Theoretical insights on the role of noise.
\newblock In \emph{The Thirteenth International Conference on Learning Representations}, 2025{\natexlab{b}}.

\bibitem[Dao et~al.(2022)Dao, Fu, Ermon, Rudra, and R{\'e}]{dao2022flashattention}
T.~Dao, D.~Fu, S.~Ermon, A.~Rudra, and C.~R{\'e}.
\newblock Flashattention: Fast and memory-efficient exact attention with io-awareness.
\newblock \emph{Advances in neural information processing systems}, 35, 2022.

\bibitem[Dosovitskiy et~al.(2020)Dosovitskiy, Beyer, Kolesnikov, Weissenborn, Zhai, Unterthiner, Dehghani, Minderer, Heigold, Gelly, et~al.]{dosovitskiy2020image}
A.~Dosovitskiy, L.~Beyer, A.~Kolesnikov, D.~Weissenborn, X.~Zhai, T.~Unterthiner, M.~Dehghani, M.~Minderer, G.~Heigold, S.~Gelly, et~al.
\newblock An image is worth 16x16 words: Transformers for image recognition at scale.
\newblock \emph{arXiv preprint arXiv:2010.11929}, 2020.

\bibitem[Goyal et~al.(2017)Goyal, Doll{\'a}r, Girshick, Noordhuis, Wesolowski, Kyrola, Tulloch, Jia, and He]{goyal2017accurate}
P.~Goyal, P.~Doll{\'a}r, R.~Girshick, P.~Noordhuis, L.~Wesolowski, A.~Kyrola, A.~Tulloch, Y.~Jia, and K.~He.
\newblock Accurate, large minibatch sgd: Training imagenet in 1 hour.
\newblock \emph{arXiv preprint arXiv:1706.02677}, 2017.

\bibitem[Grattafiori et~al.(2024)Grattafiori, Dubey, Jauhri, Pandey, Kadian, Al-Dahle, Letman, Mathur, Schelten, Vaughan, et~al.]{grattafiori2024llama}
A.~Grattafiori, A.~Dubey, A.~Jauhri, A.~Pandey, A.~Kadian, A.~Al-Dahle, A.~Letman, A.~Mathur, A.~Schelten, A.~Vaughan, et~al.
\newblock The llama 3 herd of models.
\newblock \emph{arXiv preprint arXiv:2407.21783}, 2024.

\bibitem[Hägele et~al.(2024)Hägele, Bakouch, Kosson, Allal, Werra, and Jaggi]{hagele_scaling_2024}
A.~Hägele, E.~Bakouch, A.~Kosson, L.~B. Allal, L.~V. Werra, and M.~Jaggi.
\newblock Scaling {Laws} and {Compute}-{Optimal} {Training} {Beyond} {Fixed} {Training} {Durations}, Oct. 2024.
\newblock URL \url{http://arxiv.org/abs/2405.18392}.
\newblock arXiv:2405.18392 [cs].

\bibitem[Jiang et~al.(2022)Jiang, Malik, and Li]{jiang_how_2022}
K.~Jiang, D.~Malik, and Y.~Li.
\newblock How {Does} {Adaptive} {Optimization} {Impact} {Local} {Neural} {Network} {Geometry}?, Nov. 2022.
\newblock URL \url{http://arxiv.org/abs/2211.02254}.
\newblock arXiv:2211.02254 [cs].

\bibitem[Jordan et~al.(2024)Jordan, Jin, Boza, Jiacheng, Cesista, Newhouse, and Bernstein]{jordan2024muon}
K.~Jordan, Y.~Jin, V.~Boza, Y.~Jiacheng, F.~Cesista, L.~Newhouse, and J.~Bernstein.
\newblock Muon: An optimizer for hidden layers in neural networks, 2024.
\newblock URL \url{https://kellerjordan.github.io/posts/muon/}.

\bibitem[Karpathy(2022)]{karpathy2022nanogpt}
A.~Karpathy.
\newblock Nanogpt, 2022.

\bibitem[Kingma and Ba(2014)]{kingma2014adam}
D.~P. Kingma and J.~Ba.
\newblock Adam: A method for stochastic optimization.
\newblock \emph{arXiv preprint arXiv:1412.6980}, 2014.

\bibitem[Kumar et~al.(2022)Kumar, Shen, Bubeck, and Gunasekar]{kumar2022fine}
A.~Kumar, R.~Shen, S.~Bubeck, and S.~Gunasekar.
\newblock How to fine-tune vision models with sgd.
\newblock \emph{arXiv preprint arXiv:2211.09359}, 2022.

\bibitem[Kunstner et~al.(2023)Kunstner, Chen, Lavington, and Schmidt]{kunstner_noise_2023}
F.~Kunstner, J.~Chen, J.~W. Lavington, and M.~Schmidt.
\newblock Noise is not the main factor behind the gap between sgd and adam on transformers, but sign descent might be.
\newblock In \emph{ICLR}, 2023.

\bibitem[Kunstner et~al.(2024)Kunstner, Yadav, Milligan, Schmidt, and Bietti]{kunstner_heavy-tailed_2024}
F.~Kunstner, R.~Yadav, A.~Milligan, M.~Schmidt, and A.~Bietti.
\newblock Heavy-{Tailed} {Class} {Imbalance} and {Why} {Adam} {Outperforms} {Gradient} {Descent} on {Language} {Models}, July 2024.
\newblock URL \url{http://arxiv.org/abs/2402.19449}.
\newblock arXiv:2402.19449 [cs, math, stat].

\bibitem[Liu et~al.(2024)Liu, Feng, Xue, Wang, Wu, Lu, Zhao, Deng, Zhang, Ruan, et~al.]{liu2024deepseek}
A.~Liu, B.~Feng, B.~Xue, B.~Wang, B.~Wu, C.~Lu, C.~Zhao, C.~Deng, C.~Zhang, C.~Ruan, et~al.
\newblock Deepseek-v3 technical report.
\newblock \emph{arXiv preprint arXiv:2412.19437}, 2024.

\bibitem[Liu et~al.(2025)Liu, Su, Yao, Jiang, Lai, Du, Qin, Xu, Lu, Yan, et~al.]{liu2025muon}
J.~Liu, J.~Su, X.~Yao, Z.~Jiang, G.~Lai, Y.~Du, Y.~Qin, W.~Xu, E.~Lu, J.~Yan, et~al.
\newblock Muon is scalable for {LLM} training.
\newblock \emph{arXiv preprint arXiv:2502.16982}, 2025.

\bibitem[Liu et~al.(2021)Liu, Chen, Zhou, and Zhao]{liu2018diffusion}
T.~Liu, Z.~Chen, E.~Zhou, and T.~Zhao.
\newblock A diffusion approximation theory of momentum stochastic gradient descent in nonconvex optimization.
\newblock \emph{Stochastic Systems}, 2021.

\bibitem[Loshchilov and Hutter(2016)]{loshchilov2016sgdr}
I.~Loshchilov and F.~Hutter.
\newblock Sgdr: Stochastic gradient descent with warm restarts.
\newblock \emph{arXiv preprint arXiv:1608.03983}, 2016.

\bibitem[Loshchilov and Hutter(2019)]{loshchilov2018decoupled}
I.~Loshchilov and F.~Hutter.
\newblock Decoupled weight decay regularization.
\newblock In \emph{ICLR}, 2019.

\bibitem[Malladi et~al.(2023)Malladi, Lyu, Panigrahi, and Arora]{malladi_sdes_2023}
S.~Malladi, K.~Lyu, A.~Panigrahi, and S.~Arora.
\newblock On the {SDEs} and {Scaling} {Rules} for {Adaptive} {Gradient} {Algorithms}, Feb. 2023.
\newblock URL \url{http://arxiv.org/abs/2205.10287}.
\newblock arXiv:2205.10287 [cs].

\bibitem[Mil’shtein(1986)]{mil1986weak}
G.~Mil’shtein.
\newblock Weak approximation of solutions of systems of stochastic differential equations.
\newblock \emph{Theory of Probability \& Its Applications}, 30\penalty0 (4):\penalty0 750--766, 1986.

\bibitem[Nguyen and Salazar(2019)]{nguyen2019Transformers}
T.~Q. Nguyen and J.~Salazar.
\newblock Transformers without tears: Improving the normalization of self-attention.
\newblock \emph{arXiv preprint arXiv:1910.05895}, 2019.

\bibitem[Noci et~al.(2022)Noci, Anagnostidis, Biggio, Orvieto, Singh, and Lucchi]{noci2022signal}
L.~Noci, S.~Anagnostidis, L.~Biggio, A.~Orvieto, S.~P. Singh, and A.~Lucchi.
\newblock Signal propagation in transformers: Theoretical perspectives and the role of rank collapse.
\newblock \emph{Advances in Neural Information Processing Systems}, 2022.

\bibitem[Orvieto and Gower(2025)]{orvieto2025search}
A.~Orvieto and R.~Gower.
\newblock In search of adam's secret sauce.
\newblock \emph{arXiv preprint arXiv:2505.21829}, 2025.

\bibitem[Pan and Li(2023)]{pan_toward_2023}
Y.~Pan and Y.~Li.
\newblock Toward {Understanding} {Why} {Adam} {Converges} {Faster} {Than} {SGD} for {Transformers}, May 2023.
\newblock URL \url{http://arxiv.org/abs/2306.00204}.
\newblock arXiv:2306.00204 [cs].

\bibitem[Penedo et~al.(2024)Penedo, Kydl{\'\i}{\v{c}}ek, allal, Lozhkov, Mitchell, Raffel, Werra, and Wolf]{penedo2024the}
G.~Penedo, H.~Kydl{\'\i}{\v{c}}ek, L.~B. allal, A.~Lozhkov, M.~Mitchell, C.~Raffel, L.~V. Werra, and T.~Wolf.
\newblock The fineweb datasets: Decanting the web for the finest text data at scale.
\newblock In \emph{The Thirty-eight Conference on Neural Information Processing Systems Datasets and Benchmarks Track}, 2024.
\newblock URL \url{https://openreview.net/forum?id=n6SCkn2QaG}.

\bibitem[Radford et~al.(2019)Radford, Wu, Child, Luan, Amodei, Sutskever, et~al.]{radford2019language}
A.~Radford, J.~Wu, R.~Child, D.~Luan, D.~Amodei, I.~Sutskever, et~al.
\newblock Language models are unsupervised multitask learners.
\newblock \emph{OpenAI blog}, 1\penalty0 (8):\penalty0 9, 2019.

\bibitem[Shah et~al.(2025)Shah, Polloreno, Stratos, Monk, Chaluvaraju, Hojel, Ma, Thomas, Tanwer, Shah, et~al.]{shah2025practical}
I.~Shah, A.~M. Polloreno, K.~Stratos, P.~Monk, A.~Chaluvaraju, A.~Hojel, A.~Ma, A.~Thomas, A.~Tanwer, D.~J. Shah, et~al.
\newblock Practical efficiency of muon for pretraining.
\newblock \emph{arXiv preprint arXiv:2505.02222}, 2025.

\bibitem[Shazeer(2020)]{shazeer2020glu}
N.~Shazeer.
\newblock Glu variants improve transformer.
\newblock \emph{arXiv preprint arXiv:2002.05202}, 2020.

\bibitem[Soboleva et~al.(2023)Soboleva, Al-Khateeb, Myers, Steeves, Hestness, and Dey]{cerebras2023slimpajama}
D.~Soboleva, F.~Al-Khateeb, R.~Myers, J.~R. Steeves, J.~Hestness, and N.~Dey.
\newblock {SlimPajama: A 627B token cleaned and deduplicated version of RedPajama}.
\newblock \url{https://www.cerebras.net/blog/slimpajama-a-627b-token-cleaned-and-deduplicated-version-of-redpajama}, 2023.
\newblock URL \url{https://huggingface.co/datasets/cerebras/SlimPajama-627B}.

\bibitem[Su et~al.(2024)Su, Ahmed, Lu, Pan, Bo, and Liu]{su2024roformer}
J.~Su, M.~Ahmed, Y.~Lu, S.~Pan, W.~Bo, and Y.~Liu.
\newblock Roformer: Enhanced transformer with rotary position embedding.
\newblock \emph{Neurocomputing}, 568:\penalty0 127063, 2024.

\bibitem[Tomihari and Sato(2025)]{tomihari_understanding_2025}
A.~Tomihari and I.~Sato.
\newblock Understanding {Why} {Adam} {Outperforms} {SGD}: {Gradient} {Heterogeneity} in {Transformers}, Jan. 2025.
\newblock URL \url{http://arxiv.org/abs/2502.00213}.
\newblock arXiv:2502.00213 [cs].

\bibitem[Touvron et~al.(2023)Touvron, Lavril, Izacard, Martinet, Lachaux, Lacroix, Rozi{\`e}re, Goyal, Hambro, Azhar, et~al.]{touvron2023llama}
H.~Touvron, T.~Lavril, G.~Izacard, X.~Martinet, M.-A. Lachaux, T.~Lacroix, B.~Rozi{\`e}re, N.~Goyal, E.~Hambro, F.~Azhar, et~al.
\newblock Llama: Open and efficient foundation language models.
\newblock \emph{arXiv preprint arXiv:2302.13971}, 2023.

\bibitem[Vaswani et~al.(2017)Vaswani, Shazeer, Parmar, Uszkoreit, Jones, Gomez, Kaiser, and Polosukhin]{vaswani2017attention}
A.~Vaswani, N.~Shazeer, N.~Parmar, J.~Uszkoreit, L.~Jones, A.~N. Gomez, {\L}.~Kaiser, and I.~Polosukhin.
\newblock Attention is all you need.
\newblock \emph{Advances in neural information processing systems}, 30, 2017.

\bibitem[Wang and Komatsuzaki(2022)]{wang2022gpt}
B.~Wang and A.~Komatsuzaki.
\newblock Gpt-j-6b: A 6 billion parameter autoregressive language model. 2021.
\newblock \emph{URL https://github. com/kingoflolz/mesh-transformer-jax}, page~8, 2022.

\bibitem[Wilson et~al.(2017)Wilson, Roelofs, Stern, Srebro, and Recht]{wilson2017marginal}
A.~C. Wilson, R.~Roelofs, M.~Stern, N.~Srebro, and B.~Recht.
\newblock The marginal value of adaptive gradient methods in machine learning.
\newblock \emph{Advances in neural information processing systems}, 30, 2017.

\bibitem[Xie and Li(2024)]{xie2024implicit}
S.~Xie and Z.~Li.
\newblock Implicit bias of adamw: $\ell_\infty$-norm constrained optimization.
\newblock In \emph{International Conference on Machine Learning}, pages 54488--54510. PMLR, 2024.

\bibitem[Xie et~al.(2024)Xie, Mohamadi, and Li]{xie2024adam}
S.~Xie, M.~A. Mohamadi, and Z.~Li.
\newblock Adam exploits $\ell_\infty$-geometry of loss landscape via coordinate-wise adaptivity.
\newblock \emph{arXiv preprint arXiv:2410.08198}, 2024.

\bibitem[Xiong et~al.(2020)Xiong, Yang, He, Zheng, Zheng, Xing, Zhang, Lan, Wang, and Liu]{xiong2020layer}
R.~Xiong, Y.~Yang, D.~He, K.~Zheng, S.~Zheng, C.~Xing, H.~Zhang, Y.~Lan, L.~Wang, and T.~Liu.
\newblock On layer normalization in the transformer architecture.
\newblock In \emph{International conference on machine learning}, pages 10524--10533. PMLR, 2020.

\bibitem[Zhang and Sennrich(2019)]{zhang2019root}
B.~Zhang and R.~Sennrich.
\newblock Root mean square layer normalization.
\newblock \emph{Advances in Neural Information Processing Systems}, 32, 2019.

\bibitem[Zhang et~al.(2025)Zhang, Morwani, Vyas, Wu, Zou, Ghai, Foster, and Kakade]{zhang_how_2025}
H.~Zhang, D.~Morwani, N.~Vyas, J.~Wu, D.~Zou, U.~Ghai, D.~Foster, and S.~Kakade.
\newblock How {Does} {Critical} {Batch} {Size} {Scale} in {Pre}-training?, Feb. 2025.
\newblock URL \url{http://arxiv.org/abs/2410.21676}.
\newblock arXiv:2410.21676 [cs].

\bibitem[Zhang et~al.(2020{\natexlab{a}})Zhang, He, Sra, and Jadbabaie]{zhang_why_2020-1}
J.~Zhang, T.~He, S.~Sra, and A.~Jadbabaie.
\newblock Why gradient clipping accelerates training: {A} theoretical justification for adaptivity, Feb. 2020{\natexlab{a}}.
\newblock arXiv:1905.11881 [cs, math].

\bibitem[Zhang et~al.(2020{\natexlab{b}})Zhang, Karimireddy, Veit, Kim, Reddi, Kumar, and Sra]{zhang_why_2020}
J.~Zhang, S.~P. Karimireddy, A.~Veit, S.~Kim, S.~J. Reddi, S.~Kumar, and S.~Sra.
\newblock Why are {Adaptive} {Methods} {Good} for {Attention} {Models}?, Oct. 2020{\natexlab{b}}.
\newblock URL \url{http://arxiv.org/abs/1912.03194}.
\newblock arXiv:1912.03194 [cs, math].

\bibitem[Zhang et~al.(2024)Zhang, Chen, Ding, Li, Sun, and Luo]{zhang_why_2024}
Y.~Zhang, C.~Chen, T.~Ding, Z.~Li, R.~Sun, and Z.-Q. Luo.
\newblock Why {Transformers} {Need} {Adam}: {A} {Hessian} {Perspective}, June 2024.
\newblock URL \url{http://arxiv.org/abs/2402.16788}.
\newblock arXiv:2402.16788 [cs].

\bibitem[Zhao et~al.(2024)Zhao, Morwani, Brandfonbrener, Vyas, and Kakade]{zhao_deconstructing_2024}
R.~Zhao, D.~Morwani, D.~Brandfonbrener, N.~Vyas, and S.~Kakade.
\newblock Deconstructing {What} {Makes} a {Good} {Optimizer} for {Language} {Models}, July 2024.
\newblock URL \url{http://arxiv.org/abs/2407.07972}.
\newblock arXiv:2407.07972 [cs].

\end{thebibliography}

\newpage
\appendix

{\huge Appendix}

\section{Further Experiments and Experimental Details}
\label{app:exp-details}

For pre-training Transformers on Causal Language Modeling, we use a setup that builds upon the nanoGPT~\citep{karpathy2022nanogpt} implementation,
augmenting it with Rotational Positional Embedding~\citep{su2024roformer}, RMSNorm~\citep{zhang2019root}, and SwiGLU~\citep{shazeer2020glu}. We do not adopt QK normalization or $z$-loss, as those modifications are quite recent. All our models have a vocabulary size of 50280 and make use of GPT-Neox tokenizer~\citep{black2022gpt}. We adopt an enhanced training recipe, made popular by large language models such as LLaMa~\citep{touvron2023llama}. These modifications include:
training in bfloat16; employing a linear learning rate warm-up for 10\% of the training steps~(unless specified otherwise), followed by either cosine annealing
to $1e-5$ of WSD~\citep{hagele_scaling_2024}. Global norm clipping is used~(unless specified or ablated upon) for gradients with norms above
1~(on the raw gradient, as a first step). We have no weight tying between the embedding and the last linear layer. Validation perplexity always refers to a separate subset 
consisting of 100M tokens. 

\subsection{Experimental Setup}

\paragraph{Computational Resources.} All experiments use a single NVIDIA A100-SXM4-80GB. 

\paragraph{Code.} All our runs use the repository \url{https://github.com/Niccolo-Ajroldi/plainLM}.

\paragraph{Datasets.} We test our claims on both the SlimPajama~\citep{cerebras2023slimpajama} and Fineweb~\citep{penedo2024the} datasets.

\paragraph{Model settings (12 Layers, 160M).} We use the same configuration as~\citep{biderman2023pythia}: \url{https://github.com/EleutherAI/pythia/blob/main/models/160M/pythia-160m.yml}

\begin{itemize}
    \item \textit{Layers:} 12 Transformer~\citep{vaswani2017attention} layers
    \item \textit{Attention heads:} 12
    \item \textit{Hidden size:} 768
    \item \textit{Attention implementation:} Flashattention~\citep{dao2022flashattention}.
    \item \textit{MLP type:} SwiGLU~\citep{shazeer2020glu} with expansion factor 8/3.
    \item \textit{Backbone:} PreLN Transformer~\citep{xiong2020layer} with skip connections.
    \item \textit{Normalization:} RMSnorm~\citep{zhang2019root} for both Attention and MLP.
    \item \textit{Position embeddings:} Rotary embeddings (RoPE) to 25\% of dimensions~(\citep{su2024roformer})
    \item \textit{Initialization:} the MLP and Attention output weights are initialized with variance $0.02/\sqrt{2 \# \text{layers}}$~(scaling also similar to~\citep{radford2019language}). All other weights~(comprising embeddings) are initialized with a standard deviation of $0.02$~(\citet{nguyen2019Transformers,wang2022gpt}, Sec. 2.2). Biases are always initialized at zero.
    \item \textit{Precision:} Mixed precision FP16 enabled.
    \item \textit{Dropout:} Disabled for both hidden and attention layers~(see also~\citet{chowdhery2023palm}).
\end{itemize}

\paragraph{Model settings (250 M, 24 layers).} We keep it identical to the setting above, and just increase the number of layers to 24.

\paragraph{Model settings (410 M).} We use the same setting as~\citep{biderman2023pythia}, configuration can be found here: \url{https://github.com/EleutherAI/pythia/blob/main/models/410M/pythia-410m-deduped.yml}

\begin{itemize}
    \item \textit{Layers:} 24 Transformer layers
    \item \textit{Attention heads:} 16
    \item \textit{Hidden size:} 1024
    \item Other settings as 160M parameters.
\end{itemize}

\paragraph{Model settings (1B).} We use the same setting as~\citep{biderman2023pythia}, configuration can be found here: \url{https://github.com/EleutherAI/pythia/blob/main/models/1B/pythia-1b-deduped.yml}

\begin{itemize}
    \item \textit{Layers:} 16 Transformer layers
    \item \textit{Attention heads:} 8
    \item \textit{Hidden size:} 2048
    \item Other settings as 160M parameters.
\end{itemize}

\subsection{Hyperparameter Tuning for Section~\ref{sec:big_sweeps}} Combined, the experiments in this section account for full training~(at different token budgets) of more than 250 language models at different scales and batch sizes. Every reported result is relative to the best learning rate in our grid, defined for each setup.

\paragraph{Small-scale experiments.} We consider SGD with $\beta=0.98$ and global clipping before applying momentum. For Adam, we use the setting $\beta_1=\beta_2=0.95$. Both settings are suggested by the sweep in~\autoref{fig:adam_vs_sgd_bs_sweep} and recent literature~\citep{zhang_how_2025,orvieto2025search,zhao_deconstructing_2024}.
\begin{itemize}
\item For Figure~\ref{fig:steps_tokens_bs} and Figure~\ref{fig:ppl_over_steps}~(SlimPajama, 160M), we choose a sequence length of 2048.  Inspired by the careful tuning of Figure~\ref{fig:adam_vs_sgd_bs_sweep}, we consider the learning rate grid $[0.25, 0.5, 1.0]$ for SGD and $[0.001, 0.002, 0.004]$ for Adam.
\item For Figure~\ref{fig:fw_app}~(Fineweb, 160M), we choose a sequence length of 1024. Our learning rate grid here is the same as for SlimPajama~(previous point). As a sequence length of $160k$, given our lack of experience with extremely low batch sizes~(shorter sequence length), we operate on a slightly larger grid: $[0.0001, 0.0003, 0.001, 0.003]$ for Adam and $[0.03, 0.1, 0.3, 1]$ for SGD.
\item For Figure~\ref{fig:sp_larger_app}~(SlimPajama, 250M - 24 layers), we choose a sequence length of 2048 and we also operate on a larger grid: $[0.0001, 0.0003, 0.001, 0.003]$ for Adam and $[0.03, 0.1, 0.3, 1]$ for SGD.
\end{itemize}

\begin{figure}[tbh]
  \centering
  \begin{subfigure}[t]{0.48\textwidth}
    \centering
    \includegraphics[width=\linewidth]{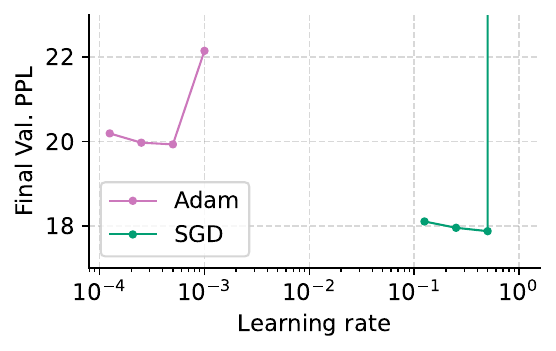}
    \caption{410M model on SlimPajama (seq. length 2048, batch size 8, 500k steps)}
    \label{fig:sweep_410M}
  \end{subfigure}
  \hfill
  \begin{subfigure}[t]{0.48\textwidth}
    \centering
    \includegraphics[width=1\linewidth]{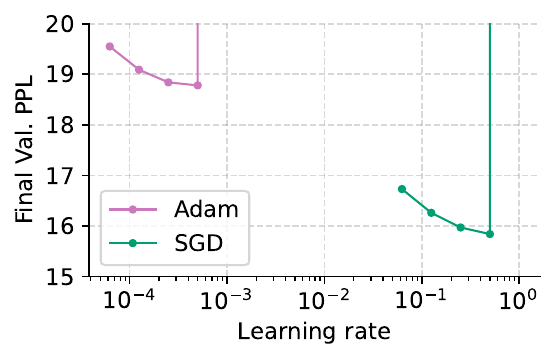}
    \caption{1B model on FineWeb (seq. length 1024, batch size 16, 850k steps)}
    \label{fig:sweep_1B}
  \end{subfigure}
  \caption{Learning rate sweep for 410M and 1B models. Trajectories for the optimal learning rate are shown in \autoref{fig:trajectory_large_models}.}
  \label{fig:sweep_large_models}
\end{figure}

\paragraph{Medium scale experiments.}
For all SGD runs, we use $\beta=0.98$. For Adam, we use the standard choice~(0.9, 0.95)~\citep{biderman2023pythia}. All our runs use global norm clipping and no weight decay.
\begin{itemize}
\item \textbf{410M model} (\autoref{fig:trajectory_410M}): We train with sequence length 2048, for $500k$ steps on SlimPajama. Learning rate grid is $[1.25e-4, 2.5e-4, 5.0e-4, 1.0e-3]$ for Adam and $[0.125, 0.25, 0.5, 1]$ for SGD. The sweep results are presented in \autoref{fig:sweep_410M}.
 \item \textbf{1B model} (\autoref{fig:trajectory_1B}): We train with sequence length 1024, for 850k steps on Fineweb. Learning rate sweep, shown in \autoref{fig:sweep_1B} uses $[6.25e-5, 1.25e-4, 2.5e-4, 5.0e-4, 1.0e-3]$ for Adam and $[0.0625, 0.125, 0.25, 0.5, 1]$ for SGD.
\end{itemize}

\section{Additional results}
\label{app:add-res}

We report the validation perplexity for the best-performing $\beta_1$ and learning rate combination for both Adam and SGD across batch sizes in~\autoref{tab:optimizer_by_bs}. The experimental setting is described in ~\autoref{sec:adamsgdgap}, and the results correspond to the sweep shown in ~\autoref{fig:adam_vs_sgd_bs_sweep}.

\begin{table}[ht]
\centering
\caption{Best validation perplexities and corresponding hyperparameters for Adam and SGD across batch sizes. Results correspond to the sweep shown in ~\autoref{fig:adam_vs_sgd_bs_sweep}.}
\vspace{2mm}
{\small
\setlength{\tabcolsep}{10pt}
\begin{tabular}{llcc}
\toprule
\textbf{Batch Size} & \textbf{Optimizer} & \textbf{PPL} & \textbf{Hyperparameters} \\
\midrule
\multirow{2}{*}{64} 
& Adam & \textbf{28.77} & $\beta_1 = 0.98,\quad \text{lr} = 1\text{e}{-}3$ \\
& SGD  & \textbf{30.76} & $\beta_1 = 0.98,\quad \text{lr} = 5\text{e}{-}1$ \\
\cmidrule(lr){1-4}
\multirow{2}{*}{256} 
& Adam & \textbf{28.20} & $\beta_1 = 0.95,\quad \text{lr} = 2\text{e}{-}3$ \\
& SGD  & \textbf{33.08} & $\beta_1 = 0.98,\quad \text{lr} = 1\text{e}{+}0$ \\
\cmidrule(lr){1-4}
\multirow{2}{*}{1024} 
& Adam & \textbf{29.36} & $\beta_1 = 0.95,\quad \text{lr} = 5\text{e}{-}3$ \\
& SGD  & \textbf{65.94} & $\beta_1 = 0.95,\quad \text{lr} = 5\text{e}{-}1$ \\
\bottomrule
\end{tabular}
}
\vspace{-2mm}
\label{tab:optimizer_by_bs}
\end{table}

In addition to \autoref{fig:ppl_over_steps}, we report the training perplexity for all other batch sizes in \autoref{fig:ppl_over_steps}. We repeat the experiments from \autoref{sec:big_sweeps} to verify that our findings generalize to a different dataset and a deeper model.

\subsection{Scaling experiments across model sizes and datasets}
\begin{figure}[tbh]
    \centering
    \includegraphics[height=0.48\linewidth]{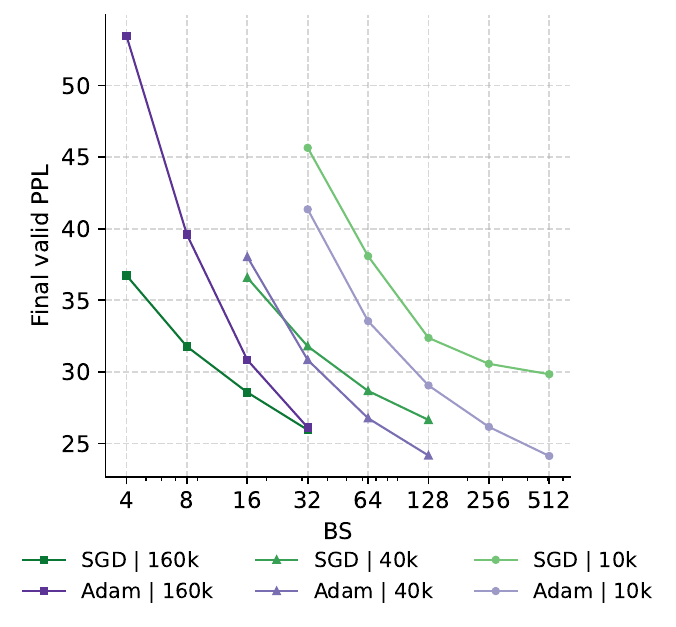}
    \caption{Fineweb dataset, sequence length 2048, 12 layers Transformer.}
    \label{fig:fw_app}
\end{figure}

First, we train the same 12-layer Transformer on the Fineweb dataset using SGD with momentum and Adam, tuning the learning rate as explained in \autoref{app:exp-details}.  Batch sizes vary from 4 to 512, and we use 3 different run lengths (i.e., different token budgets). From \autoref{fig:fw_app}, we observe that, at a fixed number of steps, the performance gap increases with batch size, and that with smaller batches and sufficiently long training, SGD outperforms Adam, consistent with the findings reported earlier.

In a second experiment, we increase the model depth to 24 layers while keeping all other settings identical to \autoref{sec:big_sweeps}. We vary batch sizes from 4 to 64 and training lengths, and tune the learning rate as explained in Appendix \ref{app:exp-details}.  As shown in Figure \autoref{fig:sp_larger_app}, the same pattern holds for a deeper model.

\begin{figure}[tbh]
    \centering
\includegraphics[width=1\linewidth]{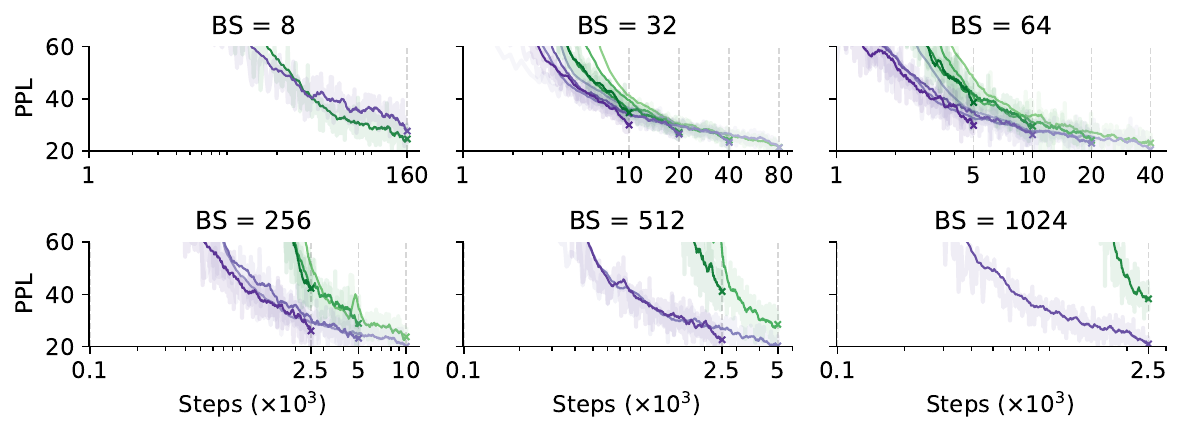}
    \caption{Perplexity during training for SGD (green) and Adam (purple) across different training lengths for all other batch sizes not shown in \autoref{fig:ppl_over_steps}. Solid lines show the rolling mean of PPL values; lighter lines show the raw values. As before, the gap decreases the longer we train, and SGD can eventually outperform Adam.
}
    \label{fig:ppl_over_steps_app}
\end{figure}


\clearpage
\newpage
    
\begin{figure}[tbh]
    \centering
    \includegraphics[height=0.48\linewidth]{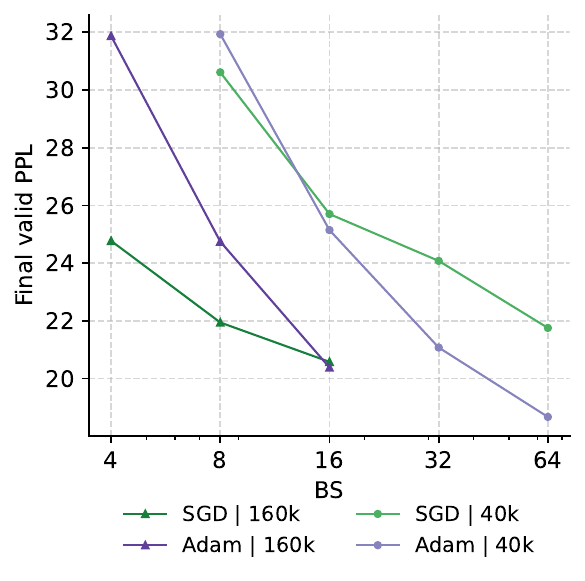}
    \caption{SlimPajama dataset, sequence length 2048, 24 layers Transformer.}
    \label{fig:sp_larger_app}
\end{figure}


\section{Algorithmic details}
\begin{algorithm}[ht]
\caption{SGD with Adaptive Momentum Clipping}
\label{alg:adapt_clip}
\begin{algorithmic}[1]
\Require Initial point $x_0$, learning rate $\eta$, momentum $\beta$, clipping fraction $p \in (0, 1)$
\For{$t = 1$ to $T$}
    \State $g_t \gets \nabla f(x_t)$ 
    \State $m_t \gets \beta m_{t-1} + g_t$
    \State Set clipping threshold $\tau_t$ as the $(1 - p)$-quantile of $|m_t|$
    \State $\hat{m}_t \gets \operatorname{clip}(m_t) = \operatorname{sign}(m_t) \cdot \min(|m_t|, \tau_t)$
    \State $x_{t+1} \gets x_t - \eta \hat{m}_t$
\EndFor
\end{algorithmic}
\end{algorithm}



\clearpage
\newpage
\section{Toy Quadratic Example}
\label{app:quadratic_details}
\begin{figure}[ht]
    \centering
    \includegraphics[width=0.99\linewidth]{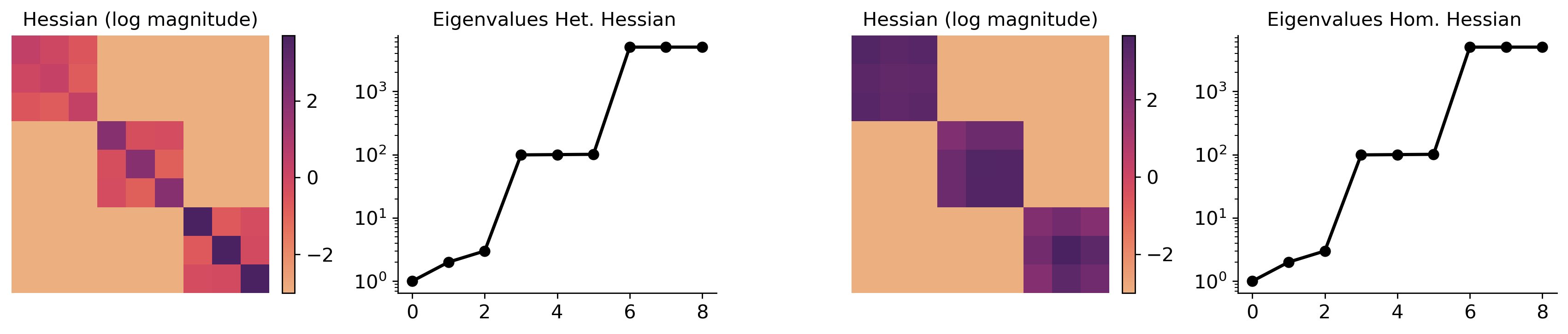}
\caption{\textit{(left) Heterogeneous and (right) Homogeneous Hessian considered in Figure~\ref{fig:quadratic}.}}
    \label{fig:quadratic-hessian}
\end{figure}

Our setup is inspired from the results and discussions in~\citet{zhang_why_2024}, and uses the codebase of~\citet{orvieto2025search}. We consider the loss

$$L(w) = \frac{1}{2}w^\top H w$$

where we construct the Homogeneous and Heterogeneous Hessians using the following procedure:
\begin{itemize}
    \item We fix the eigenvalues, equal in both cases, to
    $$\text{eig}(H_{\text{hom}}) = \text{eig}(H_{\text{het}}) =\{1,2,3,99,100,101,4998,4999,5000\}.$$
    \item We choose both Hessians to be block-diagonal, with blocks of size $3\times 3$. The homogeneous Hessian has eigenvalues of different magnitude in each block, while the Heterogeneous keeps similar magnitudes in each block.
    \begin{center}
        \verb|H_details_het = [[1,2,3],[99,100,101],[4998,4999,5000]]|\\
        \verb|H_details_hom = [[1,99,4998],[2,100,4999],[3,101,5000]]|
    \end{center}
    \item For each block, we apply a random rotation to the diagonal matrix of eigenvalues, specific to each block. Each rotation is sampled from the Haar measure by decomposition of a random $3\times 3$ positive semidefinite matrix $AA^\top$, where $A\in\mathbb{R}^{3\times 3}$ has i.i.d. Gaussian entries.
\end{itemize}
The result is shown in Figure~\ref{fig:quadratic-hessian}. Leraning rates for each method are tuned.

Next, to introduce stochasticity in this setting, we simply take the square root of the Hessian to define a $9\times 9$ design matrix $X$: $$H = X^\top X, \qquad X = H^{\frac{1}{2}},$$ and subsample a number~(the batchsize) of rows of $X$ at each iteration.

\begin{figure}[tbh]
    \centering
    \includegraphics[width=\linewidth]{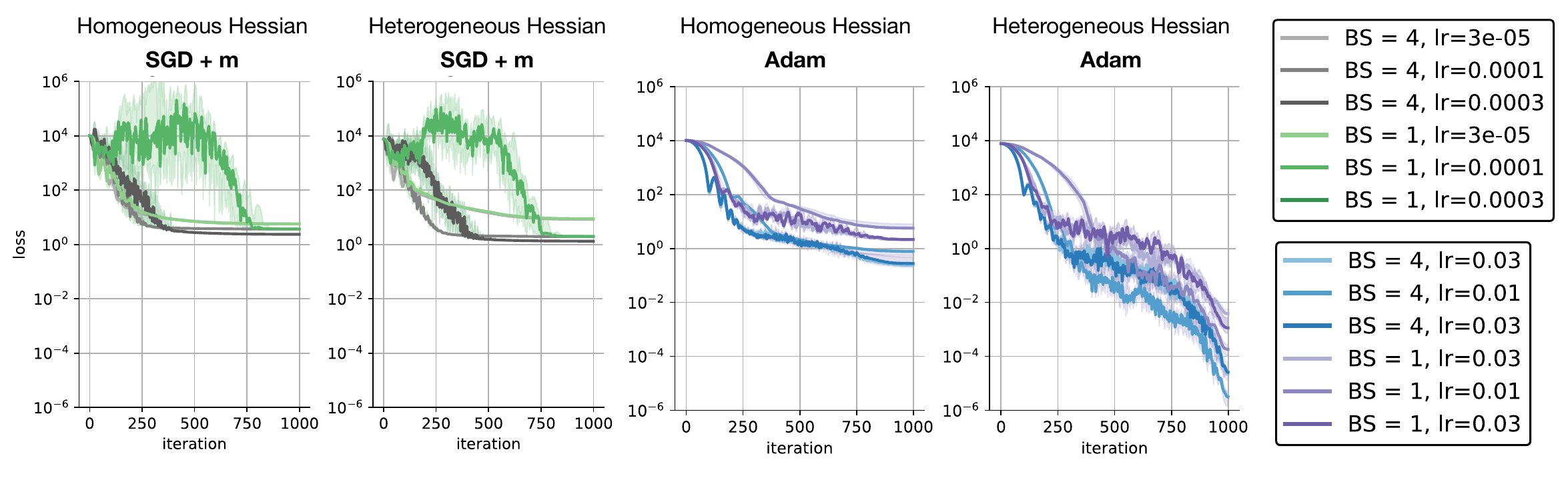}
    \caption{Complement to \autoref{fig:quadratic}.}
    \label{fig:quadratic_more}
\end{figure}

Additional learning rates for~\autoref{fig:quadratic-hessian} are reported in \autoref{fig:quadratic_more}.

\end{document}